\documentclass{amsart}

\usepackage{amssymb}
\usepackage{amsthm}

{\itshape}{\rmfamily}
{\itshape}{\rmfamily}
{\bfseries}{\itshape}
\newtheorem{definition}{Definition}{\bfseries}{\itshape}
{\itshape}{\rmfamily}
{\itshape}{\rmfamily}
\newtheorem{lemma}{Lemma}{\bfseries}{\itshape}
{\itshape}{\rmfamily}
{\itshape}{\rmfamily}
{\bfseries}{\itshape}
{\itshape}{\rmfamily}
{\itshape}{\rmfamily}
{\itshape}{\rmfamily}
\usepackage{graphicx}
\usepackage{latexsym}
\usepackage{amsmath}
\usepackage{amstext}
\usepackage{algorithm}
\usepackage[noend]{algorithmic}
\usepackage{color}
\usepackage{subfigure}
\usepackage{multirow}

\begin{document}

\title[GRASP and path-relinking for CSG]{GRASP and path-relinking for\\ Coalition Structure Generation}

\author[N. Di Mauro]{Nicola Di Mauro}
\address[Nicola Di Mauro]{Department of Computer Science, LACAM Laboratory, University of Bari
``Aldo Moro'', Bari, Italy}
\email{ndm@di.uniba.it}

\author[T.M.A. Basile]{Teresa M.A. Basile}
\address[Teresa M.A. Basile]{Department of Computer Science, LACAM Laboratory, University of Bari ``Aldo Moro'', Bari,
Italy}
\email{basile@di.uniba.it}

\author[S. Ferilli]{Stefano Ferilli}
\address[Stefano Ferilli]{Department of Computer Science, LACAM Laboratory, University of Bari ``Aldo Moro'', Bari,
Italy}
\email{ferilli@di.uniba.it}

\author[F. Esposito]{Floriana Esposito}
\address[Floriana Esposito]{Department of Computer Science, LACAM Laboratory, University of Bari ``Aldo Moro'', Bari,
Italy}
\email{esposito@di.uniba.it}

\begin{abstract}
In Artificial Intelligence with Coalition Structure Generation (CSG) one refers to those cooperative
complex problems that require to find an optimal partition, maximising a social welfare, of a set of
entities involved  in a system  into exhaustive  and disjoint coalitions.   The solution of  the CSG
problem finds applications in many fields  such as Machine Learning (covering machines, clustering),
Data   Mining  (decision   tree,  discretization),   Graph  Theory,   Natural   Language  Processing
(aggregation), Semantic Web  (service composition), and Bioinformatics.  The  problem of finding the
optimal  coalition structure is  NP-complete.  In  this paper  we present  a greedy  adaptive search
procedure  (GRASP) with  path-relinking  to efficiently  search the  space of  coalition structures.
Experiments and comparisons to other algorithms prove the validity of the proposed method in solving
this hard combinatorial problem.  
\end{abstract}

\maketitle

\section{Introduction}
\label{sec:introduction}
An active area of research in Artificial Intelligence regards methods and algorithms to solve  
complex problems that require to find an optimal partition (maximising a social welfare) of a set of
entities involved  in a system  into exhaustive  and disjoint coalitions.
This problem has been studied a lot in the area of multi-agent systems (MASs) where it is named coalition structure generation (CSG)
problem (equivalent to the complete set  partitioning problem). In particular it is interesting to
find coalition structures maximizing the sum of the values of the coalitions, that represent the
maximum payoff the agents belonging to the coalition can jointly receive by cooperating.
A coalition structure is  defined as a  partition of the  agents involved in a  system into disjoint  coalitions. The
problem of finding the optimal coalition structure  is $\mathcal{NP}$-complete~\cite{Garey90,Sandholm99}.  

Coalition generation  shares a similar  structure with a  number of common problems  in theoretical
computer science and artificial intelligence, such as in combinatorial auctions;
in job shop scheduling, Machine Learning, Data   Mining, Graph  Theory, Natural   Language
Processing, Semantic Web, and in Bioinformatics. 
In this paper we will use the term coalition structure
generation as a general term to refer to all these grouping problems. 

Sometimes there is a time limit for  finding a solution, the agents must be reactive and they
should  act as  fast as  possible.  Hence for  the specific  task  of CSG it is  necessary to  have
approximation algorithms  able to quickly  find solutions  that are within  a specific factor  of an
optimal solution. The goal of this paper is to propose a new algorithm for the CSG problem able to
quickly find a near optimal solution.

The problem of CSG has been studied in the context of characteristic function games (CFGs) in which the
value of each coalition is given by a characteristic function, and the values of a coalition structure
are obtained by summing the value of the contained coalitions. 
The  problem  of  coalition structure  generation  is  $\mathcal{NP}$-hard,  indeed as  proved
in~\cite{Sandholm99}, given $n$ the number of agents, the
number   of   possible   coalition   structures    than   can   be   generated   is   $O(n^n)$   and
$\omega(n^{n/2})$. Moreover,  in order to establish any  bound from the optimal,  any algorithm must
search at least $2^n-1$ coalition structures.  
The CSG process can be viewed as being composed of three activities~\cite{Sandholm99}:
a) \emph{coalition structure generation}, corresponding to the process of generating coalitions 
  such that agents  within each coalition coordinate their activities, but  agents do not coordinate
  between  coalitions. This  means  partitioning the  set  of agents  into  exhaustive and  disjoint
  coalitions. This partition is called a coalition structure (CS); % For instance, given four
%  agents  $\{a_1,   a_2,  a_3,   a_4\}$,  a  possible   coalition  structure   is  $\{\{a_1\}\{a_2,
%  a_3\},\{a_4\}\}$;
 b) \emph{optimization}: solving the optimization problem of each coalition.  This means pooling
  the tasks and resources of the agents in the coalition, and solving this joint problem; and 
 c) \emph{payoff distribution}: dividing the value of the generated solution among agents.
Even if these activities are independent of each other, they have some interactions. For example, the coalition that
an agent wants to join depends on the portion of the value that the agent would be allocated in each 
potential coalition. 
This paper focuses on the coalition structure generation in settings where there
are too many coalition structures to enumerate and evaluate due to costly or bounded computation and
limited time.  Instead, agents have  to select a  subset of coalition  structures on which  to focus
their search.  

In this paper we extend the work presented in~\cite{DBLP:conf/aimsa/MauroBFE10}
by adopting a stochastic local search (SLS) procedure~\cite{sls04}, named GRASP~\cite{feo95}
improved with path-relinking~\cite{Glover96tabusearch},  to solve the problem of
coalition structure generation in CFGs.  The main advantage of using a stochastic local search is to avoid
exploring an exponential number of coalition structures providing a near optimal solution. Our
algorithm  does not  provide   guarantees about  finding  the  global  optimal solution.  In
particular the questions we would like to pose are:
\textbf{Q1}) can the metaheuristic GRASP with path-relinking be used as a valuable anytime solution for the CSG
problem? In many cases, as in CSG, it is necessary to terminate the algorithm prior to completion due
to time limits and to reactivity requirements. In this situation, it is possible to adopt anytime algorithms
(i.e. algorithms that may be terminated prior to completion and returning an approximation of the correct
answer) whose quality depends on the amount of computation;
\textbf{Q2}) can  the metaheuristic  GRASP with path-relinking be  adopted for  the CSG  problem to  find optimal
  solution faster than the state of the art exact algorithms for CSG problem? In case of optimization
  combinatorial problems, stochastic  local search algorithms have been proved  to be very efficient
  in finding near optimal solution~\cite{sls04}. In many cases, they outperformed the deterministic algorithms in
  computing the optimal solution.

The  paper is  organized  as  follows: Section~\ref{sec:applications} introduces  some applications
sharing a common structure with the CSG problem, Section~\ref{sec:definitions}  presents basic  concepts
 regarding  the  CSG problem,  and  Section~\ref{sec:rw} reports  the  related  works about  the
 problem. In Sections~\ref{sec:grasp} and \ref{sec:gpr} the metaheuristic GRASP and its exension
 with path-relinking applied to the CSG problem will be
 presented. Section~\ref{sec:imp} shows some implementation details and Section~\ref{sec:conc} will conclude the paper.

\section{Applications}
\label{sec:applications}
%Many heuristic approaches have been proposed for solving combinatorial optimisation problems. 
In this section we report some common \emph{grouping problems} in theoretical computer science and
artificial intelligence that share a similar structure with the CSG problem. 

\subsection{Discretization of attributes}
An important problem in knowledge discovery and data mining is the discretization of attributes with
real values~\cite{Fayyad:1992:ASP:1867135.1867151, Chlebus:1998:FOD:646471.692492}. 
The discretization process facilitates the extraction of decision rules from a table with
real  value  attributes. As  presented  in~\cite{Chlebus:1998:FOD:646471.692492} the  discretization
selects a set of cut points of attributes  determining a partition of the real value attributes into
intervals.  The set of  cuts determines  a grid  in $k$-dimensional  space with  $\prod_{i=1}^k n_i$
regions, where $k$  is the number of attributes and  $n_i$ is the number of  intervals of the $i$-th
attribute. Checking if there is a consistent set of cuts such that the grid defined by them contains
at most $K$ regions in NP-complete~\cite{Chlebus:1998:FOD:646471.692492}.

\subsection{Learning Bayesian prototype trees}
In \cite{myllymaki95} the authors present a method to learn Bayesian networks, and specifically
Bayesian prototype trees assuming that data form clusters of similar vectors. The data can be 
partitioned into clusters and the maximum likelihood estimates of the partitioning computed. A data
partition determines the corresponding Bayesian prototype tree model. Hence, given a training data
set $\mathcal D$ consisting of $N$ data vectors, the authors proposed to find the optimal Bayesian
prototype tree by finding the optimal partition vector among the  $N^N$ different vectors.

\subsection{Cluster ensemble problem with graph partitioning}
Clustering and graph partitioning are two concepts strongly related. Clustering is a data analysis technique adopted in statistics, data mining, and machine learning
communities, involving partitioning a set of instances into a given number of groups
optimising an objective function. Recently, \emph{cluster ensemble}
techniques~\cite{Strehl:2003:CEK:944919.944935} improve clustering
performance by generating multiple partitions of the given data set and then combining them to form
a superior clustering solution. \cite{Strehl:2003:CEK:944919.944935,DBLP:conf/icml/FernB04} propose
a graph partitioning formulation for cluster ensembles. Given a data set $x = \{X_i, \ldots, X_n\}$,
a cluster ensemble is a set of clustering solutions, represented as $C = \{C_1, \ldots, C_R\}$,
where $R$ is the ensemble size. Each clustering solution $C_i$ is a partition of the data set into
$K_i$ disjoint clusters.
The graph partitioning problem is to partition a weighted graph $G$ into $K$ parts by finding $K$
disjoint clusters of its vertices. $G$ is characterised by the set $V$ of vertices and by a
nonnegative and symmetric similarity matrix $W$ characterising the similarity between each pair of
vertices. The cut of a partition $P$ is defined as $Cut(P,W)=\sum w(i,j)$,
where $i$ and $j$ are vertices do not belonging to the same cluster. The general goal of graph
partitioning is to find a partition that maximise the cut.

\subsection{Aggregation for Natural Language Generation}
\emph{Aggregation} represents a main component of natural language generation systems. The task is to merge
two or more linguistic structures into a single sentence. In~\cite{barzilay-a} the authors presented an
automatic tool for performing the semantic grouping task by formalising it as a CSG problem, where
each coalition corresponds to a sentence. The strength of the proposed approach lies in its ability
to capture global partitioning constraints by performing collective inference over local pairwise
assignments. Pairwise constraints capture the semantic compatibility between pairs of linguistic
structures at local level. The global task is to search a semantic grouping that maximally agrees
with the pairwise preferences while simultaneously satisfies constraints on the partitioning as a
whole.

\subsection{Privacy-preserving data mining}
Privacy-preserving data  mining~\cite{Verykios:2004:SPP:974121.974131} is a new research area that is focused  on preventing
privacy violations  that might arise during  data mining operations.  To reach this goal many
techniques modify original datasets in  order to preserve privacy even  after the mining
process is activated ensuring minimal data loss and obtaining qualitative data mining results. 
In~\cite{Matatov:2010:PDM:1801027.1801464} the authors presented a technique considering
anonymisations for classification through feature set partitioning. Feature set partitioning
decomposes the original set of features into many subsets in order to induce a classifier for each
subset. Then unlabelled instances are classified combining the votes of all classifiers.
The work in~\cite{Matatov:2010:PDM:1801027.1801464} solves the problem of preserving k-anonymity via
feature set partitioning as follows. Given a learner $I$, a combination method $C$, and a training
set $S$ with input feature set $A$, the goal is to find an optimal partitioning of the input feature
set $A$ into $w$ mutually exclusive subsets. Optimality is defined in terms of minimisation
of the generalization error of the classifier combined using the method $C$.

\section{Definitions}
\label{sec:definitions}
In this section we present the basic notions about the CSG problem.
\begin{definition}[Coalition]
Given a set $N = \{a_1, a_2, \ldots, a_n\}$ of $n$ agents, $|N|=n$, called the \emph{grand coalition}, a
\emph{coalition} $S$ is a non-empty subset of the set $N$, $\emptyset \neq S \subseteq N$.  
\end{definition}

\begin{definition}[Coalition Structure]
  A \emph{coalition structure} (CS), or \emph{collection}, $\mathcal C = \{C_1, C_2, \ldots, C_k \}
  \subseteq 2^N$ is a partition of
the set $N$, and $k$ is  its size, i.e. $\forall i,j: C_i  \cap C_j = \emptyset$ and $\cup_{i=1}^k
C_i = N$. 
\end{definition}
Given $\mathcal C = \{C_1, C_2, \ldots, C_k \}$, we define  $\cup \mathcal C \triangleq \cup_{i=1}^k C_i$. We
will denote the set of all coalition structures of $N$ as $\mathcal M(N)$. 

As reported
in~\cite{RePEc:wsi:igtrxx:v:11:y:2009:i:03:p:347-367}, assuming a comparison relation
$\triangleright$, $\mathcal A \triangleright \mathcal B$ means that the way $\mathcal A$ partitions
$N$, where  $N = \cup \mathcal A = \cup \mathcal B$, is preferable to the way $\mathcal B$
partitions $K$.

In this paper we consider the following rules that allow us to trasform partitions of the grand
coalition. Given a  CS $\mathcal C = \{C_1, C_2, \ldots, C_t\}$: 
\begin{description}
\item[SPLIT:] $C \rightarrow C \setminus \{C_i\} \cup \{C_k,C_h\}$, where
  $C_k \cup C_h = C_i$, with $C_k, C_h \neq \emptyset$;
\item[MERGE:]  $C \rightarrow C  \setminus \{C_i,C_j\}_{i \neq
    j} \cup \{C_k\}$, where $C_k = C_i \cup C_j$;
\item[SHIFT:] $C \rightarrow C  \setminus \{C_i,C_j\}_{i \neq
    j} \cup \{C_i',  C_j'\}$, where $C_i' = C_i  \setminus \{a_i\}$ and $C_j' = C_j  \cup
  \{a_i\}$, with $a_i \in C_i$.
\end{description}

As in common practice~\cite{Sandholm99,Rahwan08}, we consider coalition structure generation in \emph{characteristic function games}
(CFGs). A CFG is a pair $(N,v)$, where $N=\{a_1,a_2,\ldots,a_n\}$ and $v$ is a function 
 $v :  2^N \rightarrow \mathbb R$. Given $\mathcal C$ a coalition structure, $v(\mathcal C) = \sum_{C_i \in \mathcal C} v(C_i)$, 
where  $v(C_i)$ is the value of the  coalition $C_i$.  Intuitively,
$v(C_i)$ represents the maximum payoff the members of $C_i$ can jointly receive by cooperating.  As 
in~\cite{Sandholm99}, we assume  that $v(C_i) \geq 0$. In  case of negative values, it  is possible to
normalize the coalition values, obtaining a game strategically equivalent to the original game~\cite{Kahan84}, by subtracting a lower bound value from all coalition values.
For CFGs the comparison relation on coalition structures is induced in a canonical way,
$\mathcal A \triangleright \mathcal B \iff v(\mathcal A) \triangleright v(\mathcal B)$.

It is possible to prove the following results:
\begin{lemma}[\cite{keinanen09}]
Given two coalition structures $\mathcal A, \mathcal B \in \mathcal M(N)$, with $\mathcal A \neq
\mathcal B$, then $\mathcal A$ can be transformed into $\mathcal B$ by doing at most $n-1$
applications of the SPLIT or the MERGE rules.
\end{lemma}

\begin{lemma}[\cite{keinanen09}]
Given two coalition structures $\mathcal A, \mathcal B \in \mathcal M(N)$, with $\mathcal A \neq
\mathcal B$, then $\mathcal A$ can be transformed into $\mathcal B$ by doing at most $n-1$
applications of the SHIFT rule.
\end{lemma}

Now we can define the coalition structure problem as follows.
\begin{definition}[Coalition structure generation problem]
  Given a set of agents $A$, the \emph{coalition structure generation} problem is to
maximize the social welfare of the agents by finding a coalition structure 
$C^* = \arg\max_{C \in \mathcal M(A)} v(C)$.
\end{definition}

Formally, a CSG problem may be formulated as a set partitioning problems (SPP). 
Let $I = \{1, \ldots, m\}$ be a set of objects, and let $\{P_1, \ldots, P_n\}$ be a collection of
subsets of $I$, with a cost $c_j \in \mathbb R^+$ associated with each subset $P_j$.
Given a $n \times m$ binary matrix $A = \{a_{ij}\}$, where $a_{ij}=1$ if $i\ in P_j$ and $a_{ij}=0$
otherwise, let $\widehat{J}$ be a solution of SPP represented as the $n$-dimensional vector $\vec{x}
= \langle x_i, \ldots, x_n \rangle$ of binary decision variables. An integer programming formulation
of the set partitioning problem is
$$z(x) = \max \sum_{j=1}^n c_j x_j$$
$$\mathrm{subject\ to} \ \sum_{j=1}^n a_{ij}x_j = 1, i=1\ldots m.$$

Given $n$ agents, the size of the input to a CSG algorithm is exponential,
since  it contains the  values $v(\cdot)$  associated to  each of  the $(2^n-1)$  possible coalitions.
Furthermore,  the  number of  coalition  structures  grows as  the  number  of  agents  increases  and
corresponds to $\sum_{i=1}^n Z(n,i)$, where $Z(n,i)$, also known as the Stirling number of the
second kind, is  the number of coalition structures  with $i$ coalitions, and may  be computed using
the following recurrence: $Z(n,i) = i Z(n-1,i) + Z(n-1,i-1)$, where $Z(n,n) = Z(n,1) = 1$. 
As   proved   in~\cite{Sandholm99},  the   number   of   coalition   structures  is   $O(n^n)$   and
$\omega(n^{n/2})$, and hence an exhaustive enumeration becomes prohibitive. 

In this  paper we focus  on games that  are neither \emph{superadditive} nor  \emph{subadditive} for
which  the  problem  of coalition  structure  generation  is  computationally complex.  Indeed,  for
superadditive games  where $v(S \cup T)  \geq v(S) + v(T)$  (meaning any two  disjoint coalitions are
better off by merging together), and for subadditive games  where $v(S \cup T) < v(S) + v(T)$ for all
disjoint  coalitions  $S,T   \subseteq  A$,  the  problem  of   coalition  structure generation  is
trivial.  In particular,  in  superadditive  games, the  agents  are better  off  forming the  grand
coalition where all agents operate together ($C^*  = \{A\}$), while in subadditive games, the agents
are better off by operating alone ($C^* = \{ \{a_1\}, \{a_2\}, \ldots, \{a_n\}\}$).

Instances  of the  CSG  problem have  been defined  using the  following
distributions, as proposed in~\cite{Larson:1999:ACS:301136.301158,Rahwan09}, for the values of the characteristic function $v$: 
\begin{itemize}
  \item  Uniform (U): $v(C) \sim U(a,b)$ where $a=0$ and $b=1$;
  \item  Uniform Scaled (US): $v(C) \sim |C| \cdot U(a,b)$ where $a=0$ and $b=1$;
  \item Normal (N): $v(C) \sim N(\mu,\sigma^2)$ where $\mu =1$ and $\sigma=0.1$; 
  \item Normal Scaled (NS): $v(C) \sim |C| \cdot N(\mu,\sigma^2)$ where $\mu =1$ and $\sigma=0.1$; 
  \item Normally Distributed (ND): $v(C) \sim N(\mu,\sigma^2)$ where $\mu =|C|$ and
    $\sigma=\sqrt{|C|}$.
\end{itemize}

\begin{figure}
  \centering
  \subfigure[Uniform]{\includegraphics[width = 0.49\textwidth]{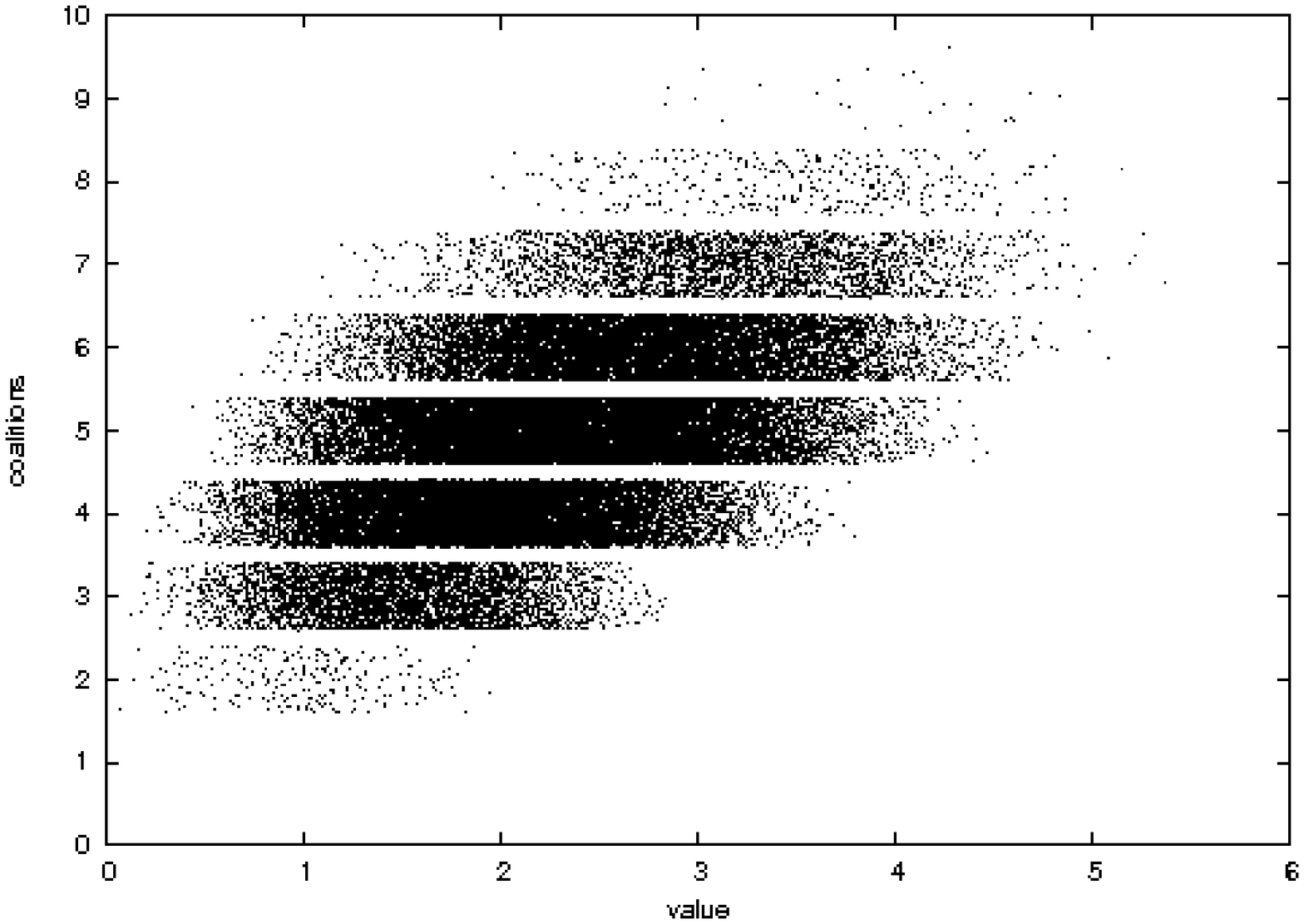}}
  \subfigure[Uniform scaled]{\includegraphics[width = 0.49\textwidth]{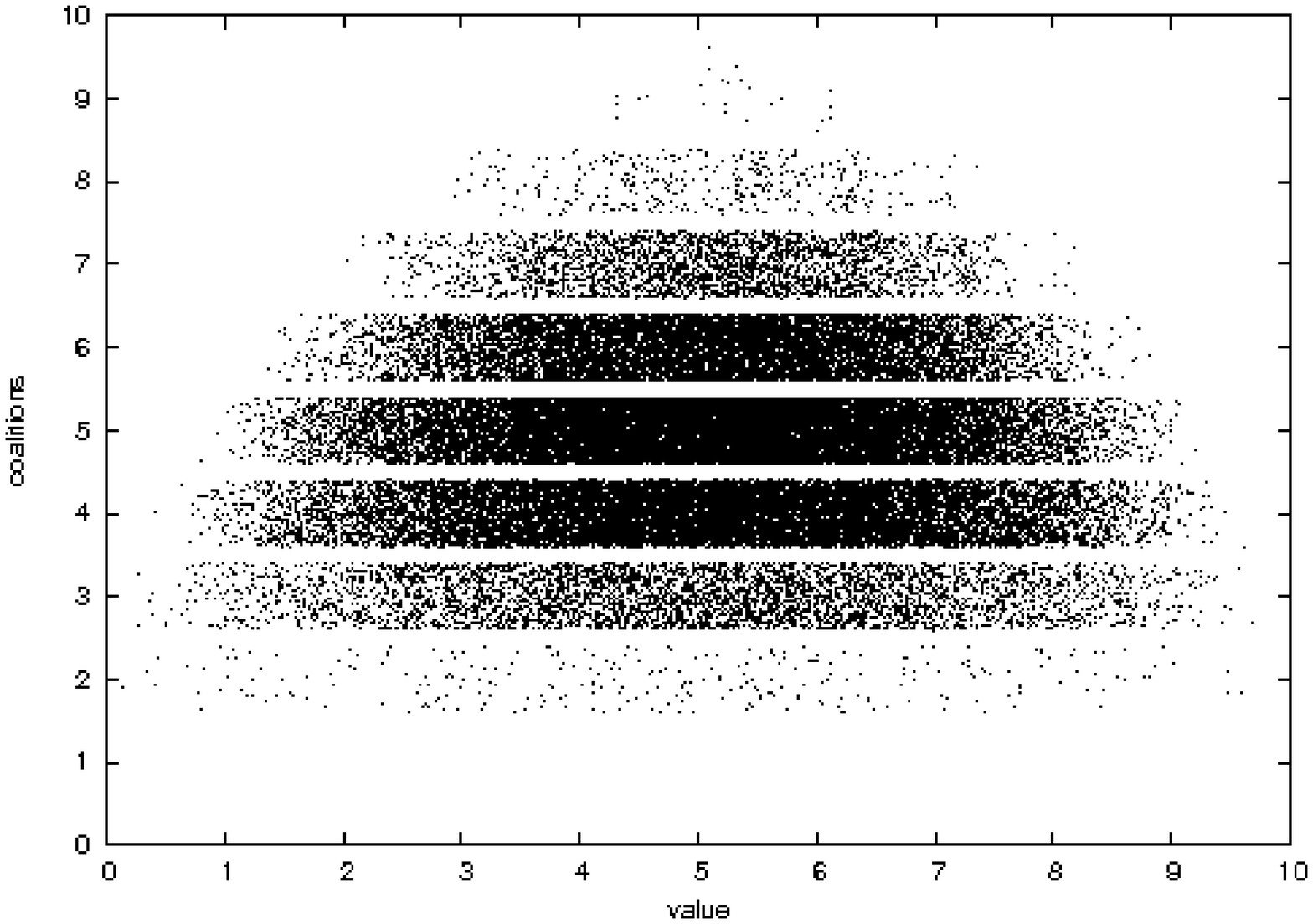}}
  \subfigure[Normal]{\includegraphics[width = 0.49\textwidth]{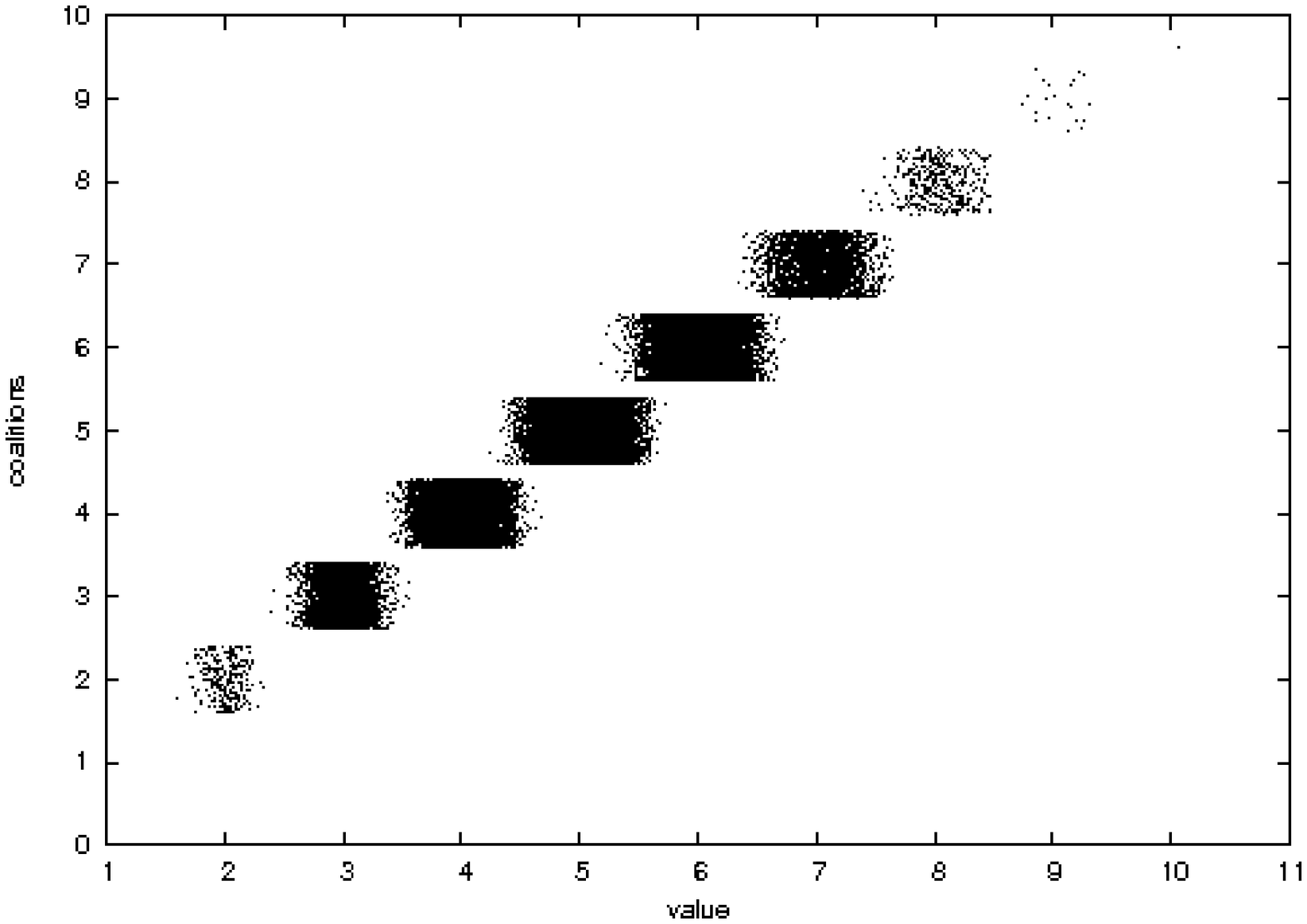}}
  \subfigure[Normal scaled]{\includegraphics[width = 0.49\textwidth]{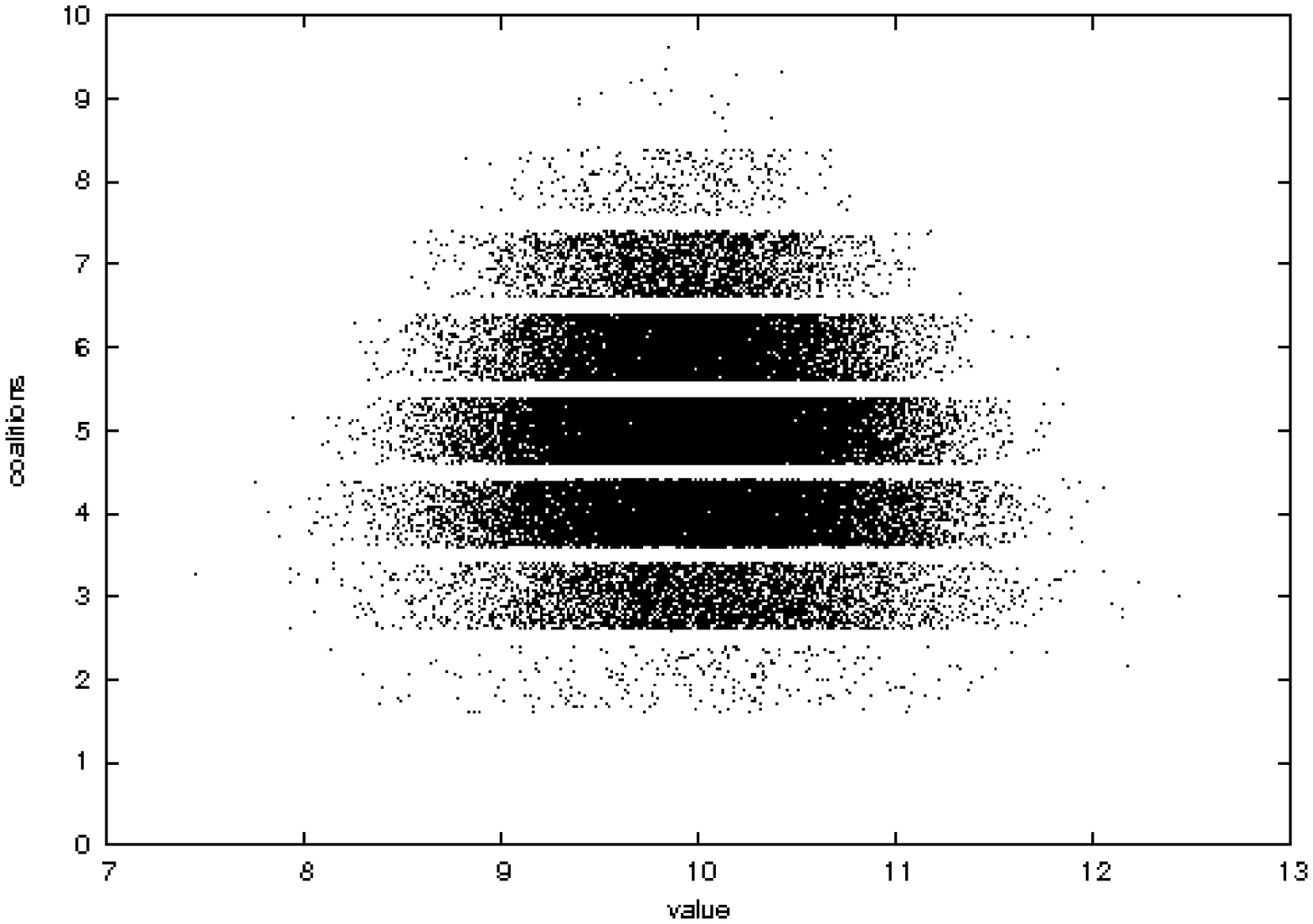}}
  \subfigure[Normally distributed]{\includegraphics[width = 0.49\textwidth]{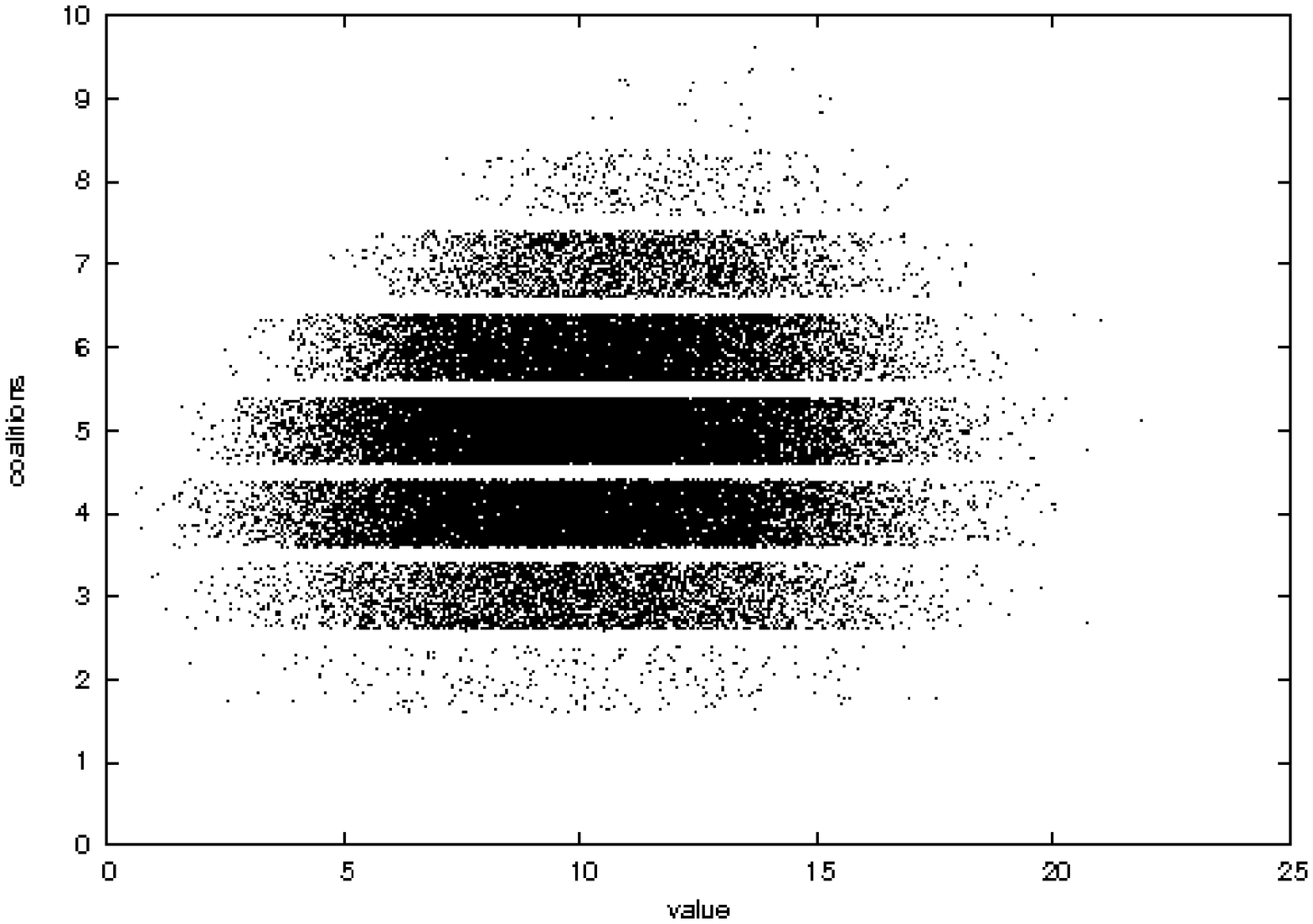}}
  \caption{Plots  of the coalition structures' values according to the Uniform,
  Uniform scaled, Normal, Normal  scaled, and Normally
  Distributed distributions of the characteristic function for 10 agents.} 
    \label{fig:distr}
\end{figure}

Figure~\ref{fig:distr} plots the coalition structures' values according to the previous five
distributions of the characteristic function for 10 agents. Each graph plots on the x-axis the value
of the coalition structures whose cardinality is represented by a point on the y-axis.  As we can see, it seems to be easy to find
optimal solutions in the case of normal and uniform distributions, while it becomes more complicated
for the case of scaled distributions. For the normal distribution, the optimal
solution belongs to the less populated region corresponding to CSs with none or ten coalitions. The
same scenario arises for the uniform distribution, even if here the region containing the optimal
solution is more populated than in the previous case. The CFGs with scaled distributions are very
hard to solve since the optimal solution may belong to very populated regions.
More formally, given $n$ agents, let $\mathbf{c}$ be the random variable of the value of a CS with $k$
coalitions $c_i$. For each distribution the expected value of the $\mathbf{c}$ variable may be computed as follows:
$\mathbb{E}_{U}(\mathbf{c}) = \sum_i^k \mathbb{E}(c_i) = k/2$ (maximum when $k =
n$, many coalitions); $\mathbb{E}_{US}(\mathbf{c}) = \sum_i^k |c_i|/2$ (maximum with few
coalitions where it is more probable to assign a high value to each one); 
$\mathbb{E}_{N}(\mathbf{c}) = \sum_i^k 1 = k$ (maximum when $k=n$);
$\mathbb{E}_{NS}(\mathbf{c}) = \sum_i^k |C_i| $ (maximum with few coalitions);
$\mathbb{E}_{ND}(\mathbf{c}) = \sum_i^k |C_i| $ (maximum with few coalitions).

%Ad ogni passo del processo di costruzione data  una $CS = \{ c_1, \ldots, c_k\}$ posso aggiungere il
%nuovo agente  ad una delle  coalizioni $c_i$ oppure  come nuova. Nel  secondo caso il  valore sar\`a
%sicuramente migliore. Nel caso della uniforme questa strategia mi porta ad avere strutture con molte
%coalizioni. Nel caso uniforme scalato invece costruiremo strutture con poche coalizioni.
%
%Se divido il valore per il numero di coalizioni:

\section{Related Work}
\label{sec:rw}
Previous works  on CSG can  be broadly divided into  two main categories:
exact algorithms that  return an optimal solution,  and approximate algorithms that  find an approximate
solution with limited resources.

A deterministic algorithm must systematically explore the search space of candidate solutions. 
One of the first  algorithms returning an optimal solution is the  dynamic programming algorithm (DP)
proposed in~\cite{Yeh86} for  the set partitioning problem.%\footnote{The set  partitioning problem is
% equivalent to the coalition structure generation problem.}. 
This algorithm is  polynomial in the
size of  the input  ($2^n-1$)  and it  runs in $O(3^n)$  time, which is  significantly less  than an
exhaustive enumeration ($O(n^n)$). However, DP is not an anytime algorithm, and has a large memory
requirement.  Indeed, for each coalition $C$ it computes the tables $t_1(C)$ and $t_2(C)$. It computes all the
possible splits  of the  coalition $C$ and  assigns to $t_1(C)$  the best  split and to  $t_2(C)$ its
value. 
In~\cite{Rahwan08} the  authors proposed an  improved version of  the DP algorithm  (IDP) performing
fewer operations and requiring less memory than DP.  IDP, as shown by the authors, is considered one
of the fastest available exact algorithm in the literature computing an optimal solution.

\begin{figure}
  \centering
  \includegraphics[width = 0.8\textwidth]{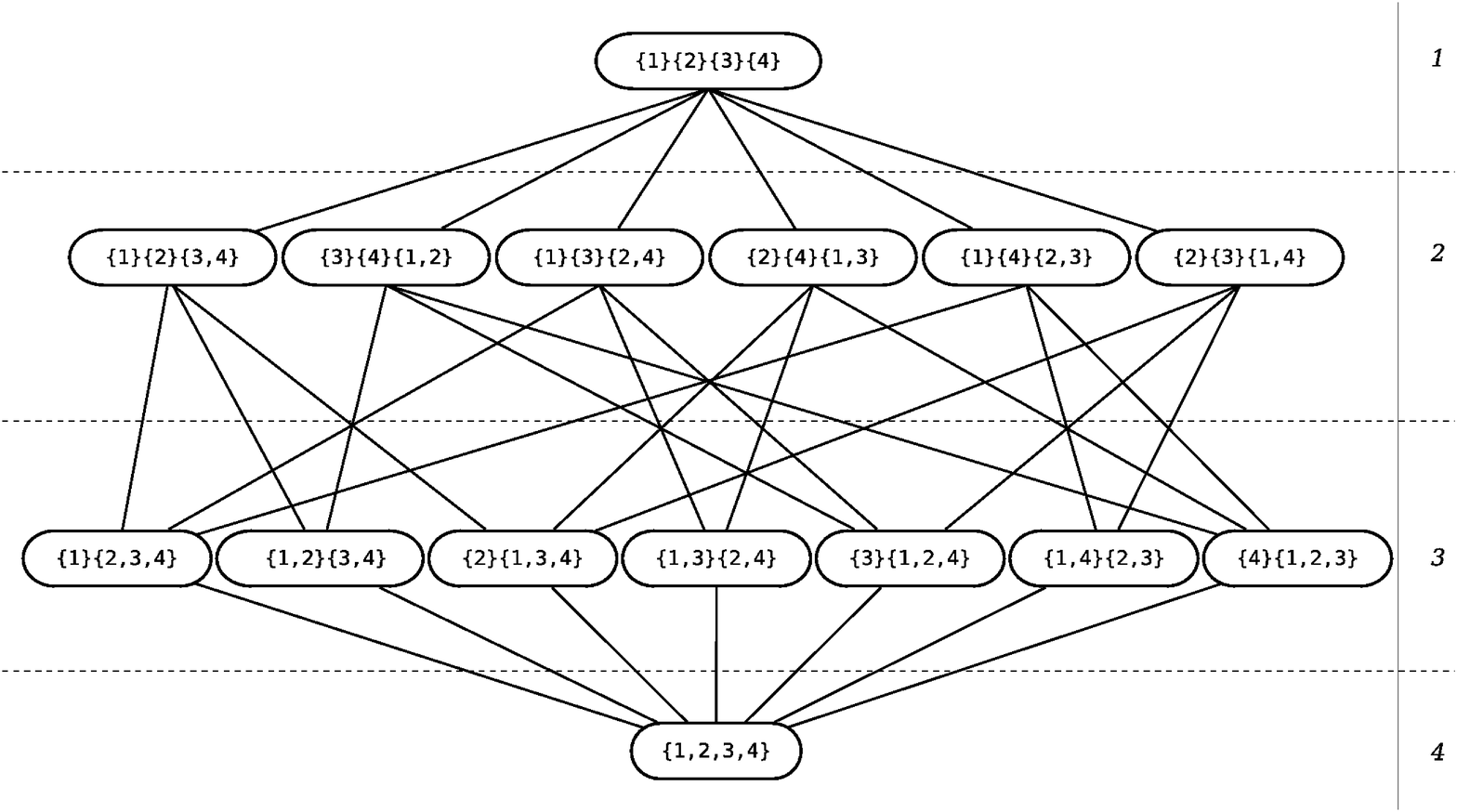}
  \caption{Coalition structure graph for a 4-agent game.}
  \label{sandholmgraph}
\end{figure}

Given a coalition $C$, $|C|=n$, the number of splitting of $C$ into
two coalitions  $C_1$ and $C_2$,  with $|C_1|=s_1$ and  $|C_2|=s_2$ is
computed as follows

\begin{equation*}
S(s_1,s_2) = \left\{ \begin{array}{ll}
  \mathbf C(s_1+s_2,s_2) / 2 &\mbox{ if $s_1=s_2$} \\
  \mathbf C(s_1+s_2,s_2)  &\mbox{ otherwise}
        \end{array} \right.
\end{equation*}

 where  $\mathbf C(n,k) =  {n\choose k}$  is the  binomial coefficient,
 i.e. the number of $k$-combinations from a set with $n$ elements.
 Now, the total number of splittings computed by DP is 
\begin{equation*}
  \mathbf  S_{DP}  = \sum_{s=1}^n  \mathbf  C(n,s) \sum_{k=\lceil  s/2
    \rceil}^{s-1} S(s-k,k)
 \end{equation*}
 while those computed by IDP is
 \begin{equation*}
   \mathbf S_{IDP} = \sum_{s=1}^n \mathbf C(n,s) \left ( \sum_{k=\lceil s/2
     \rceil}^{s-1}  S(s-k,k) \mathbf  1_{\{k  \leq n  -  s \vee  s=n\}}
 \right )
 \end{equation*}
 where $\mathbf 1_{\{k \leq n - s \vee s=n\}}$ is $1$ if $k \leq n - s$ or
 $s=n$, $0$ otherwise.

Both  DP  and IDP  are  not  anytime  algorithms, they  cannot  be  interrupted before  their  normal
termination.   In~\cite{Sandholm99},  Sandholm  et  al.   have   presented  the  first  anytime
algorithm, sketched in Algorithm~\ref{alg:sandholm}, 
that can  be interrupted to obtain a solution  within a time limit but  not guaranteed to be
optimal.  When not interrupted it returns the optimal solution. 
The CSG process can be  viewed as a search in a coalition structure graph
as reported in Figure~\ref{sandholmgraph}.
One desideratum is to be able to guarantee that the CS is within a worst case bound 
from optimal, i.e. that searching through a subset $N$ of coalition structures,
$k = \min \{k' \} \ \mathrm{where} \ k' \geq \frac{V(S^*)}{V(S^*_N)}$ 
is finite, and as small as possible, where $S^*$ is  the best CS and $S^*_N$ is the best CS that has
been seen in the subset $N$.
In~\cite{Sandholm99} has  been proved that: a)  to bound $k$, it  suffices to search  the lowest two
levels of the coalition structure graph (with this  search, the bound $k=n$, and the number of nodes
searched is $2^{n-1}$); b) this bound is tight;  and, c) no other search algorithm can establish any
bound $k$ while searching only $2^{n-1}$ nodes or fewer.

\begin{algorithm}
\caption{Sandholm et al. algorithm}
\label{alg:sandholm}
\begin{enumerate}
\item Search the bottom two levels of the coalition structures graph.
\item Continue with a breadth-first search from the top of the graph as long as there is time left,
or until  the entire graph has  been searched (this  occurs when this breadth-first  search completes
level 3 of the graph, i.e. depth n-3).
\item Return the coalition structure that has the highest welfare among those seen so far.
\end{enumerate}
\end{algorithm}

A new anytime algorithm has been proposed in~\cite{Rahwan09}, named IP, whose idea is to partition the space
of the possible solutions into  sub-spaces such that it is possible to compute  upper and lower bounds on
the values of the  best CSs they contain. Then, these bounds  are used to prune all
the sub-spaces that cannot contain the  optimal solution. Finally, the algorithm searches through the
remaining sub-spaces  adopting a  branch-and-bound technique avoiding  to examine all  the solutions
within the searched sub-spaces. IP can be used to find optimal coalition structure avoiding to
search most of the search space. As reported in ~\cite{Rahwan09}, IP finds optimal solutions much
faster than any previous algorithm designed for this purpose.

As regards the approximate algorithms, in~\cite{sen00} it has been proposed a solution based on a genetic
algorithm, which performs well when there is some regularity in the search space. Indeed, the authors
assume, in  order to  apply their algorithm,  that the value  of a  coalition is dependent  of other
coalitions in the CS, making the algorithm not well suited for the general case.
A  new  solution~\cite{keinanen09} is  based  on  a Simulated  Annealing  algorithm~\cite{Kirkpatrick83},  a widely  used
stochastic local search method. At each iteration the algorithm  selects a random neighbour solution
$s'$ of a CS $s$. The search proceeds with an adjacent CS $s'$ of the original CS $s$ if $s'$ yields
a better  social welfare than  $s$. Otherwise,  the search is  continued with $s'$  with probability
$e^{(V(s')-V(s))/t}$,  where  $t$  is the  temperature  parameter  that  decreases according  to  the
annealing schedule $t = \alpha t$.

\section{GRASP for the CSG problem}
\label{sec:grasp}
The resource limits posed by some intelligent systems, such as the time for finding a solution, require  to have
approximation algorithms  able to quickly  find solutions  that are within  a specific factor  of an
optimal solution.  
In this  section we firstly present  the anytime  algorithm for CSG proposed
in~\cite{DBLP:conf/aimsa/MauroBFE10} with some improvements and then its extension with path-relinking.

A  method to  find  high-quality solutions  for  a combinatorial  problem is  a  two steps  approach
consisting of a  greedy construction phase followed by  a perturbative\footnote{A perturbative local
  search  changes  candidate solutions  by  modifying  one or  more  of  the corresponding  solution
  components.} local search~\cite{sls04}. Namely, the greedy  construction  method  starts the process from  an  empty candidate  solution and at each  construction step  adds the  best ranked
component  according  to  a   heuristic  selection  function. Successively,  a perturbative  local search algorithm is used  to improve the
candidate  solution  thus  obtained. Advantages  of  this
search method, over other stochastic local search algorithms, are the much better solution quality and fewer
perturbative  improvement steps  to  reach the  local optimum.  Greedy
Randomized Adaptive  Search Procedures (GRASP)~\cite{feo95}  solve the
problem  of  the  limited  number  of  different  candidate  solutions
generated by  a greedy construction search methods  by randomising the
construction  method.  GRASP  is an  iterative  process,  in  which  each  iteration consists  of  a
construction phase, producing a feasible solution, and a local search 
phase, finding a local optimum  in the neighborhood of the constructed
solution.        The       best       overall        solution       is
returned.

\begin{algorithm}[tb]
\caption{GRASP CSG}
\label{alg:grasp}
\begin{algorithmic}[1]
\REQUIRE{$v$: the characteristic function;\\
$A$: the set of $n$ agents;\\
\texttt{maxIter}: maximum number of iterations; \\
\texttt{neighOp}: neighbourhood operator; \\
\texttt{riiSteps}: max non improving search steps for the RII procedure; \\
\texttt{wp}: RII walk probability}
\ENSURE{solution $\widehat{C} \in \mathcal M(A)$}
\STATE $\widehat{C} = \emptyset$, $v(\widehat{C}) = -\infty$
\STATE iter $= 0$
\WHILE{iter $<$ \texttt{maxIter}}
  \STATE $\alpha = $ rand(0,1); 
  \STATE $C = \emptyset$; $i = 0$
  \STATE \emph{/* construction */}
  \WHILE{$i < n$}   
    \STATE $\mathcal S = \{ C' | C' = add(C,A)\}$
    \STATE $\overline{s} = \max \{v(T) | T \in \mathcal C\}$
    \STATE $\underline{s} = \min \{v(T) | T \in \mathcal C\}$
    \STATE RCL $= \{C' \in \mathcal S | v(C') \geq \underline{s} + \alpha (\overline{s} - \underline{s})\}$
    \STATE randomly select an element $C$ from RCL
    \STATE $i \leftarrow i + 1$
  \ENDWHILE
  \STATE \emph{/* local search */}
  \STATE $C = \texttt{RandomisedIterativeImprovement}(C, \texttt{wp}, \texttt{riiSteps},
  \texttt{neighOP})$
%  \STATE $\mathcal N = \{ C' \in neigh(C) | v(C') > v(C)\}$
%  \WHILE{$\mathcal N \neq \emptyset$}
%    \STATE select $C \in \mathcal N$
%    \STATE $\mathcal N \leftarrow \{ C' \in neigh(C) | v(C') > v(C)\}$
%  \ENDWHILE
  \IF{$v(C) > v(\widehat{C})$}
  \STATE $\widehat{C} = C$
  \ENDIF
  \STATE iter = iter + 1
\ENDWHILE
\STATE \textbf{return} $\widehat{C}$
\end{algorithmic}
\end{algorithm}

Algorithm~\ref{alg:grasp}  reports   the  outline  of  the GRASP procedure for the CSG problem,
denoted in the following with \texttt{GRASP}.  
In  each  iteration,  it
computes a solution $C$ by using a randomised constructive search procedure and then applies a local
search  procedure to  $C$ yielding  an improved  solution. The  main procedure  is made up of two
components: a constructive phase (lines 7-13) and a local search phase (line 15). 
The constructive search algorithm used in \texttt{GRASP} iteratively adds a solution component by randomly
selecting it,  according to  a uniform  distribution, from a  set, named  \emph{restricted candidate
  list}  (RCL), of  highly ranked  solution components  with respect  to a  greedy function  $g  : C
  \rightarrow \mathbb R$.  The probabilistic component  of \texttt{GRASP} is characterized by randomly choosing
one of  the best candidates  in the  RCL.  In our  case the greedy  function $g$ corresponds  to the
characteristic function $v$ presented in Section~\ref{sec:definitions}.  In particular, given $v$, the heuristic function, and $\mathcal C$, the set of feasible
solution components, $\underline{s} = \min \{ v(C) | C \in \mathcal C\}$ and $\overline{s} = \max \{
v(C)  | C  \in \mathcal  C\}$ are  computed. Then  the RCL  is defined  by including  in it  all the
components $C$  such that  $v(C) \geq  \underline{s} + \alpha  (\overline{s} -  \underline{s})$. The
parameter  $\alpha$  controls the  amounts  of  greediness and  randomness.  A  value  $\alpha =  1$
corresponds to a greedy construction procedure, while $\alpha = 0$ produces a random construction. 
As  reported  in~\cite{Mockus97},  GRASP  with  a  fixed  nonzero  RCL  parameter  $\alpha$  is  not
asymptotically convergent  to a global optimum.   The solution to make  the algorithm asymptotically
globally convergent, could be to randomly  select the parameter value from the continuous interval
$[0, 1]$ at the beginning of each iteration  and using this value during the entire iteration, as we
implemented in \texttt{GRASP}.

Given a set of nonempty subsets of $n$ agents $A$, $C = \{C_1, C_2,
\ldots,  C_t\}$, such  that $C_i  \cap C_j  \neq \emptyset$  and $\cup  C \subset  A$,  the function
$add(C,A)$ used in the construction phase returns a refinement $C'$ obtained from $C$ using
one of the following operators:
\begin{enumerate}
\item  $C' \rightarrow C \setminus \{C_i\} \cup
\{C'_i\}$ where $C'_i = C_i \cup \{a_i\}$ and $a_i \not\in \cup C$, or 
\item $C' \rightarrow C \cup
\{C_i\}$ where $C_i = \{a_i\}$ and $a_i \not\in \cup C$.
\end{enumerate}
Starting from the empty set, in the first iteration all the  coalitions containing exactly one agent
are  considered and  the best  is selected  for further  specialization. At  the iteration  $i$, the
working set of coalition $C$ is refined by trying to add an agent to one of the coalitions in $C$ or
a new coalition containing the new agent is added to $C$.

\subsection{Local search: Randomised Iterative Improvement}

To improve  the solution generated by  the construction phase, a  local search is used.  It works by
iteratively replacing the current solution with a better solution taken from the neighborhood of the
current solution while there is a better solution in the neighborhood.
In order  to build  the neighborhood of  a coalition  structure $C$ we adopted the previously
reported operators  SPLIT, MERGE and SHIFT, leading to the following two
neighbourhood relations: 
\begin{itemize}
  \item $N_{s/m}(\mathcal C) = \{ C' \in S | s' \in SPLIT(s) \cup MERGE(s)\}$
  \item $N_{s}(\mathcal C) = \{ \mathcal C' \in S | s' \in SHIFT(s) \}$
\end{itemize}

In particular, as a local search procedure in \texttt{GRASP} we used a Randomised Iterative Improvement (RII)
technique~\cite{sls04}, 
as reported in Algorithm~\ref{rii}. The algorithm starts from the solutions obtained in the
constructive phase of \texttt{GRASP}, and then tries to improve the current  candidate solution with respect
to $v$. RII uses a parameter $wp \in [0,1]$, called walk probability, that corresponds to the
probability of performing a random walk step (line 12) instead of an improvement step (lines 14-23).
The uninformed random walk randomly selects a solution from the complete neighbourhood (line 12).
The improvement step randomly selects one of the strictly improving neighbours $I(\mathcal C)$ (line
16) or a minimally worsening neighbour if the set $I(\mathcal C)$ is empty (line 18).
The search is terminated when a given number of search steps (\texttt{steps}) has been performed without achieving
any improvement (line 4). 

\begin{algorithm}[tb]
\caption{Randomised Iterative Improvement (RII)}
\label{alg:rii}
\begin{algorithmic}[1]
  \REQUIRE{$\mathcal{C}$: candidate solution;\\
  \texttt{neighOP}: neighbourhood operator; \\
  \texttt{steps}: max non improving search steps; \\
  \texttt{wp}: walk probability}
  \ENSURE{$\widehat{\mathcal{C}}$: candidate solution}
  \STATE improvingSteps = 0
  \STATE bestValue = $v(\mathcal C)$
  \STATE $\widehat{\mathcal C} = \mathcal C$
  \WHILE{improvingSteps $<$ \texttt{steps}}
  \STATE improvingSteps++
\IF{\texttt{neighOP} == SPLIT/MERGE} 
  \STATE compute the neighborhooud $N(\mathcal C)=N_{s/m}(\mathcal C)$ of $\mathcal C$ 
\ELSE
  \STATE compute the neighborhooud $N(\mathcal C)=N_{s}(\mathcal C)$ of $\mathcal C$ 
\ENDIF
\STATE $u$ = rand([0,1])
\IF{$u \geq \texttt{wp}$} 
%\STATE \COMMENT{$x^* = $ step$_{urw}(\pi,x)$, uninformed random walk}
\STATE randomly select $\mathcal C'$ from $N(\mathcal C)$
\ELSE
%\STATE \COMMENT{$x^* = $ step$_{ii}(\pi,x)$, iterative improvement}
\STATE $I(\mathcal C) = \{\mathcal S \in N(\mathcal C) | v(\mathcal S) < v(\mathcal C)\}$ 
\IF{$I(\mathcal C)\neq \emptyset$} 
\STATE randomly select $\mathcal C'$ from $I(\mathcal{C})$
\ELSE
\STATE select $\mathcal C'$ from $N(\mathcal C)$  such that $\forall \mathcal S \in N(\mathcal C): v(\mathcal{C'}) < v(\mathcal S)$
\ENDIF
\ENDIF
\STATE $\mathcal C = \mathcal C'$
\IF{$v(\mathcal C) < $ bestValue}
\STATE bestValue $= v(\mathcal C)$
\STATE improvingSteps = 0
\STATE $\widehat{\mathcal C} = \mathcal C$
\ENDIF
\ENDWHILE
\STATE \textbf{return} $\widehat{\mathcal C}$

\end{algorithmic}
\label{rii}
\end{algorithm}

\subsection{GRASP evaluation}
\label{sec:grasp_eval}

Stochastic Local Search algorithms are typically incomplete when applied to a  given instance of an
optimisation problem and the time required for finding a solution may be considered as a random
variable~\cite{sls04}.
Given an  optimisation SLS algorithm $S$  for an optimisation  problem $\Pi$ and a  soluble instance
$\pi  \in \Pi$,  let $P(T_{S,\pi}  \leq t,  Q_{S,\pi} \leq  q)$ denote  the probability  that $S$
applied to $\pi$ finds a solution of quality less than or equal to $q$ in time less than or equal to
$t$. The \emph{run-time distribution}  (RTD) of $S$ on the specific instance $\pi$ is the  probability distribution of the
bivariate  random  variable  $(T_{S,\pi},  Q_{S,\pi})$,  characterised  by  the  \emph{run-time
distribution function} $rtd: \mathbb R^+ \times \mathbb R^+ \rightarrow [0,1]$ defined as $rtd(t,q) =
P(T_{S,\pi} \leq t, Q_{S,\pi} \leq q)$~\cite{sls04}.
To empirically measure RTDs, let $k$ be the  total number of runs performed with a cutoff time $t'$,
and  let $k'  < k$  be the  number of  successful runs (i.e., runs during  which a  solution was
found). Let $rt(j)$  denote the run-time for the $j$th entry  in the list of successful
runs, ordered according to increasing run-times. The cumulative empirical RTD is then defined by 
$\hat{P}(T \leq t) := \# \{j | rt(j) \leq t\}/k$.

A CPU time measurement is always based  on specific implementations and run-time environments. It is often more appropriate 
to measure run-time in a way that allows one to abstract from these factors. This can be 
done using \emph{operation counts}, reflecting the number of operations that are
considered to contribute significantly towards an algorithm's performance. Run-time measurements
corresponding to actual CPU times and abstract run-times measured in operation
counts may be distinguished by referring to  the latter as \emph{run-lengths}. We refer  to RTDs obtained from run-times
measured in terms of operation counts as \emph{run-length distributions} or RLDs~\cite{sls04}.

In order to evaluate the proposed algorithms, we implemented them in the C language and the corresponding source code has been
included  in   the  ELK  system\footnote{ELK is a system  including many algorithms for the CSG problem  whose source code is
  publicly available at \texttt{http://www.di.uniba.it/$\sim$ndm/elk/}.}.
ELK includes also our implementation of the  algorithm proposed by Sandholm et al. in \cite{Sandholm99}, DP~\cite{Yeh86},
IDP~\cite{Rahwan08}, and IP~\cite{Rahwan09}. 

All the following experimental  results about the  behaviour of both \texttt{GRASP} and its
extension with path-relinking are obtained executing the algorithms included in ELK on PC with an Intel(R) Core(TM) i5 CPU
670 @ 3.47GHz and 8GB of RAM, running GNU/Linux kernel 2.6.32-25-server.

\begin{figure}
  \centering
  \subfigure[Uniform]{\includegraphics[width = 0.49\textwidth]{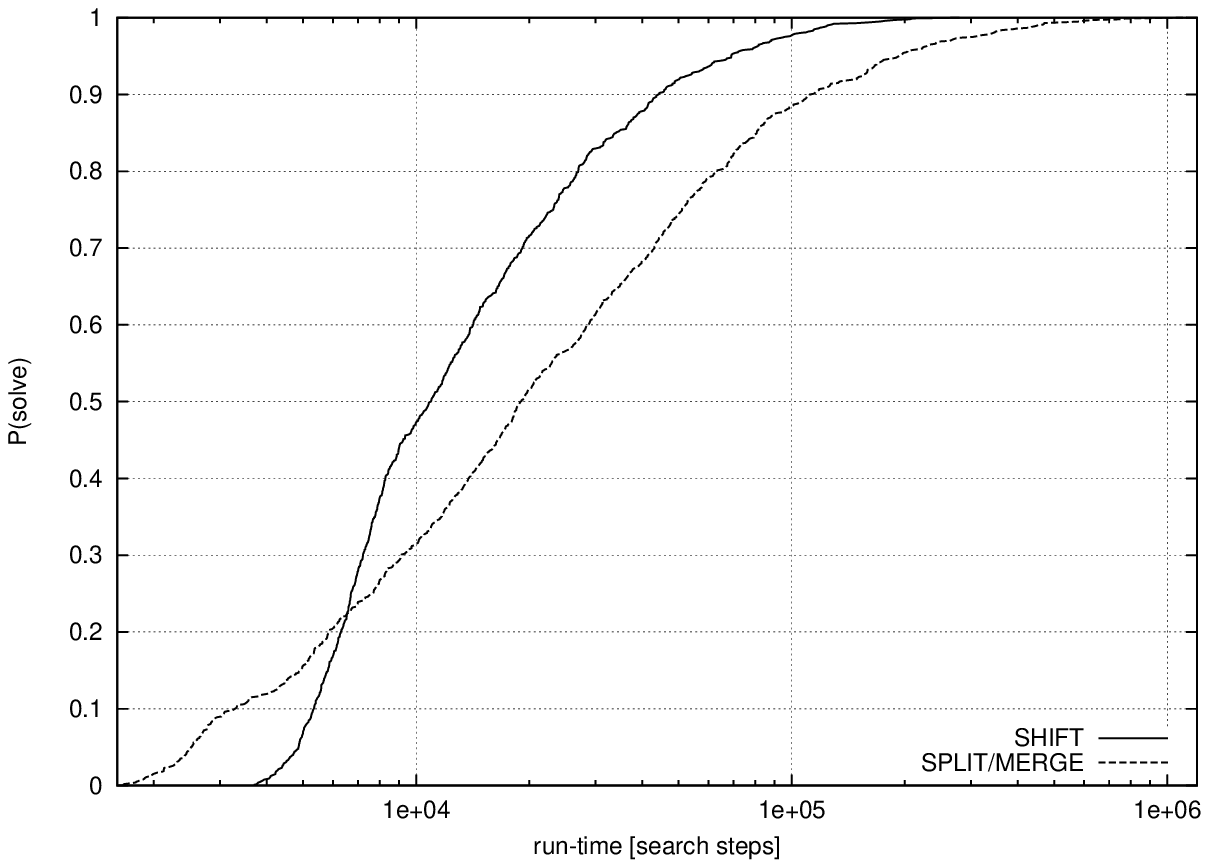}}
  \subfigure[Uniform scaled]{\includegraphics[width = 0.49\textwidth]{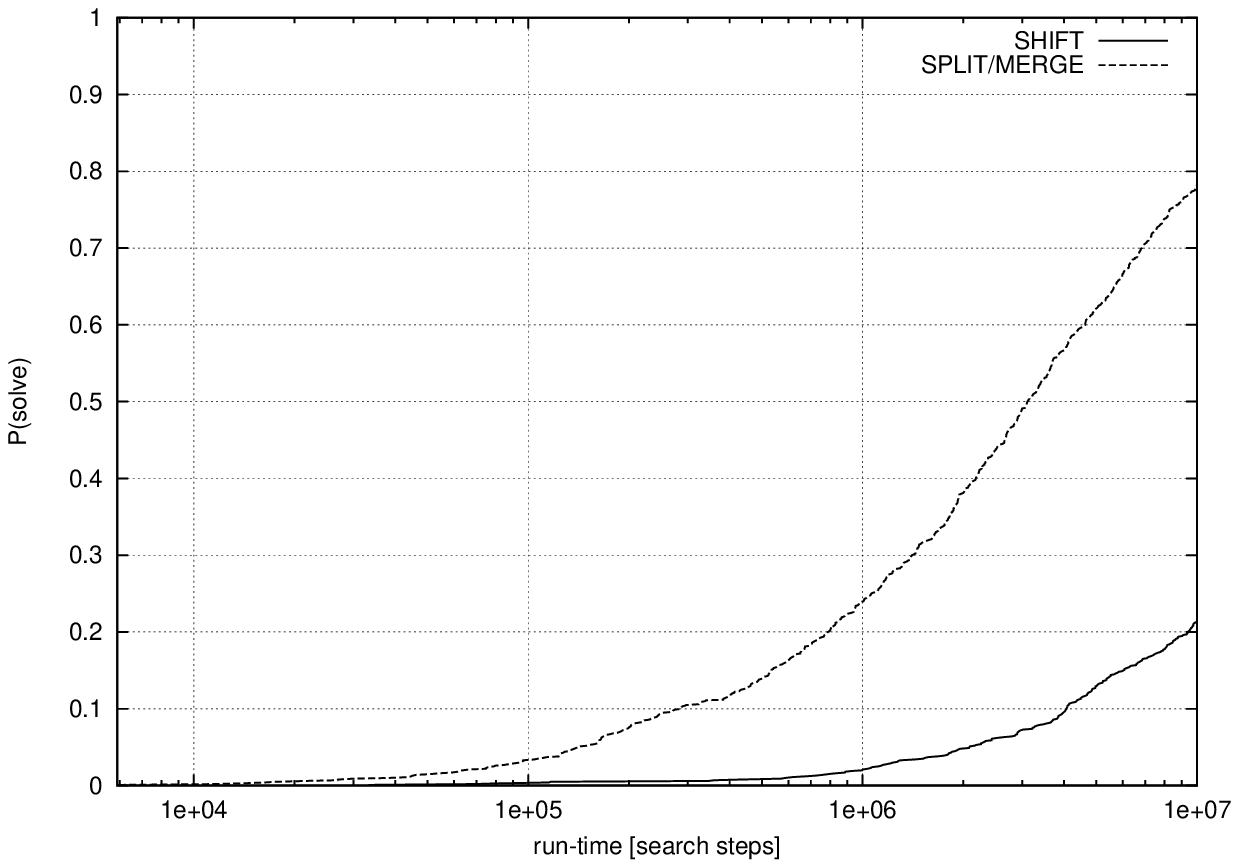}}
  \subfigure[Normal]{\includegraphics[width = 0.49\textwidth]{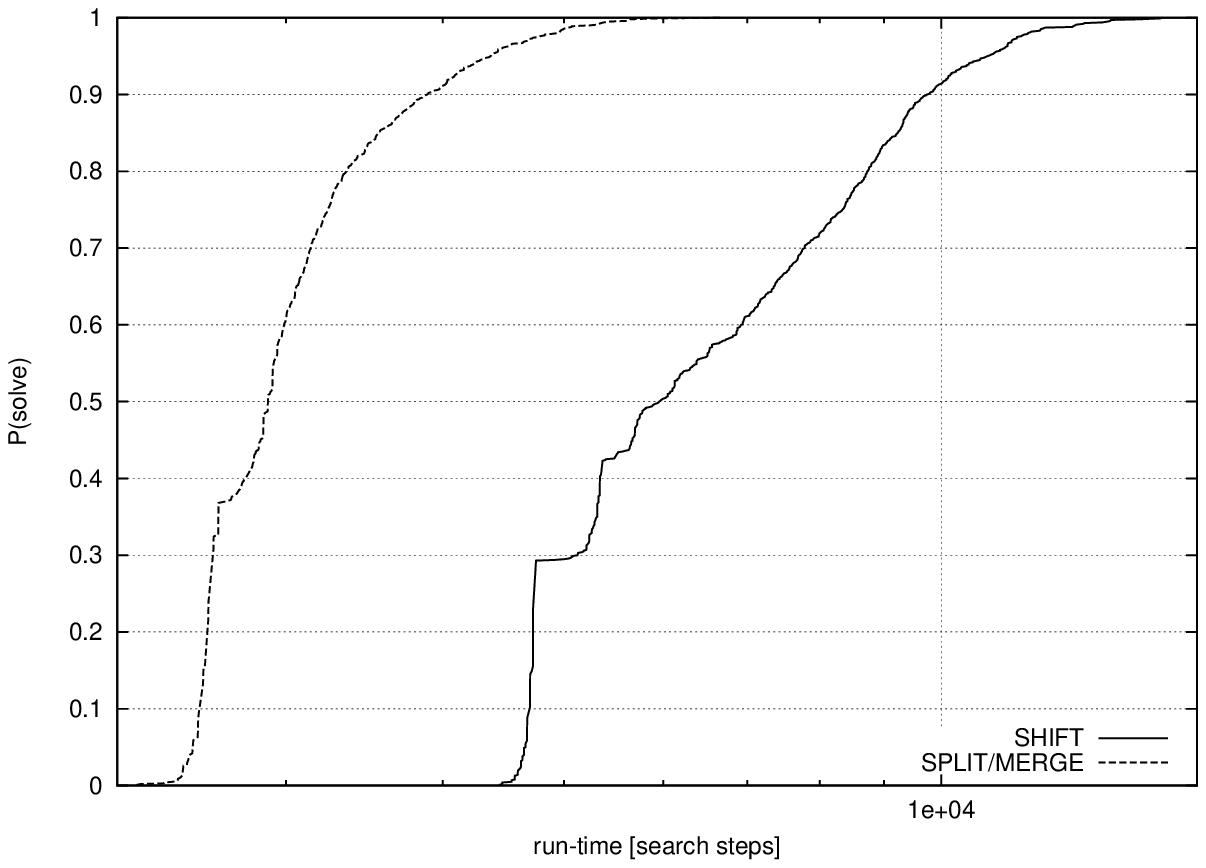}}
  \subfigure[Normal scaled]{\includegraphics[width = 0.49\textwidth]{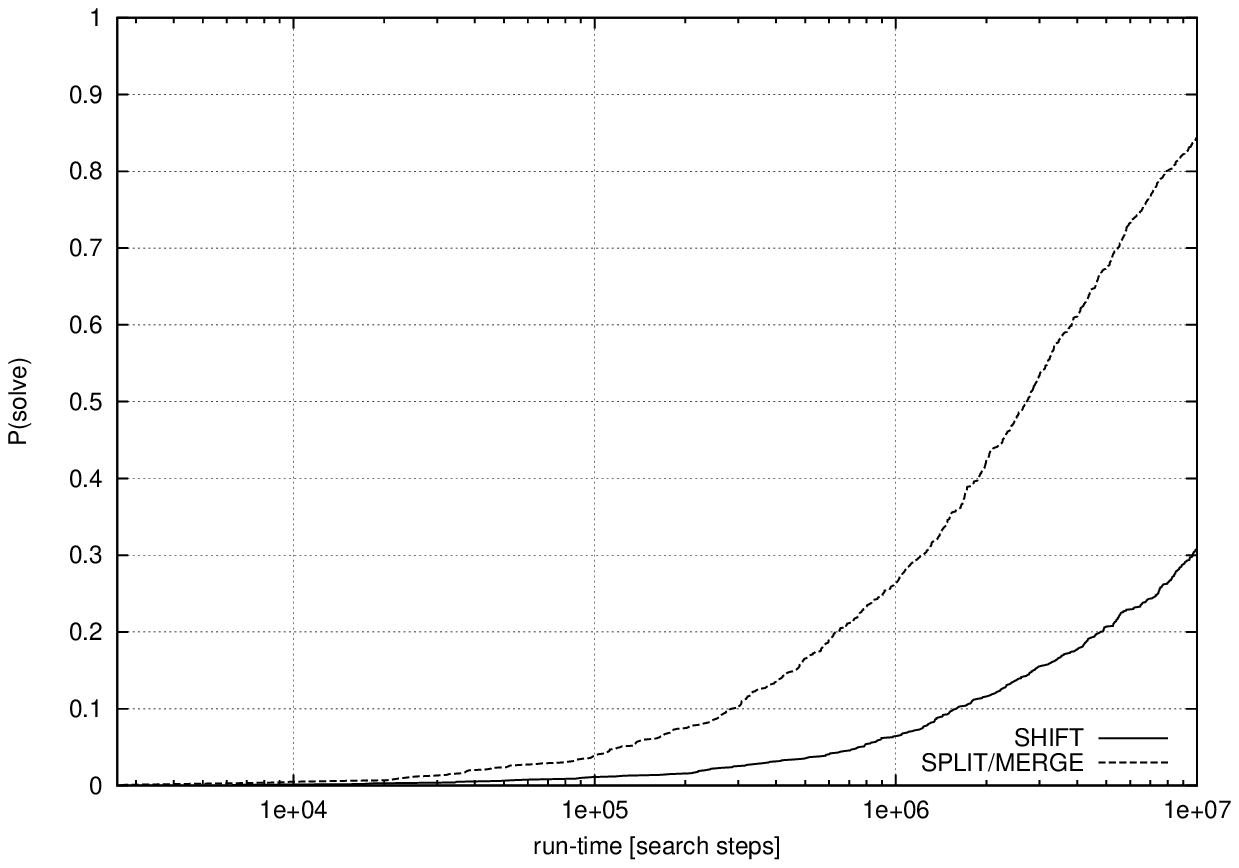}}
  \subfigure[Normal distributed]{\includegraphics[width = 0.49\textwidth]{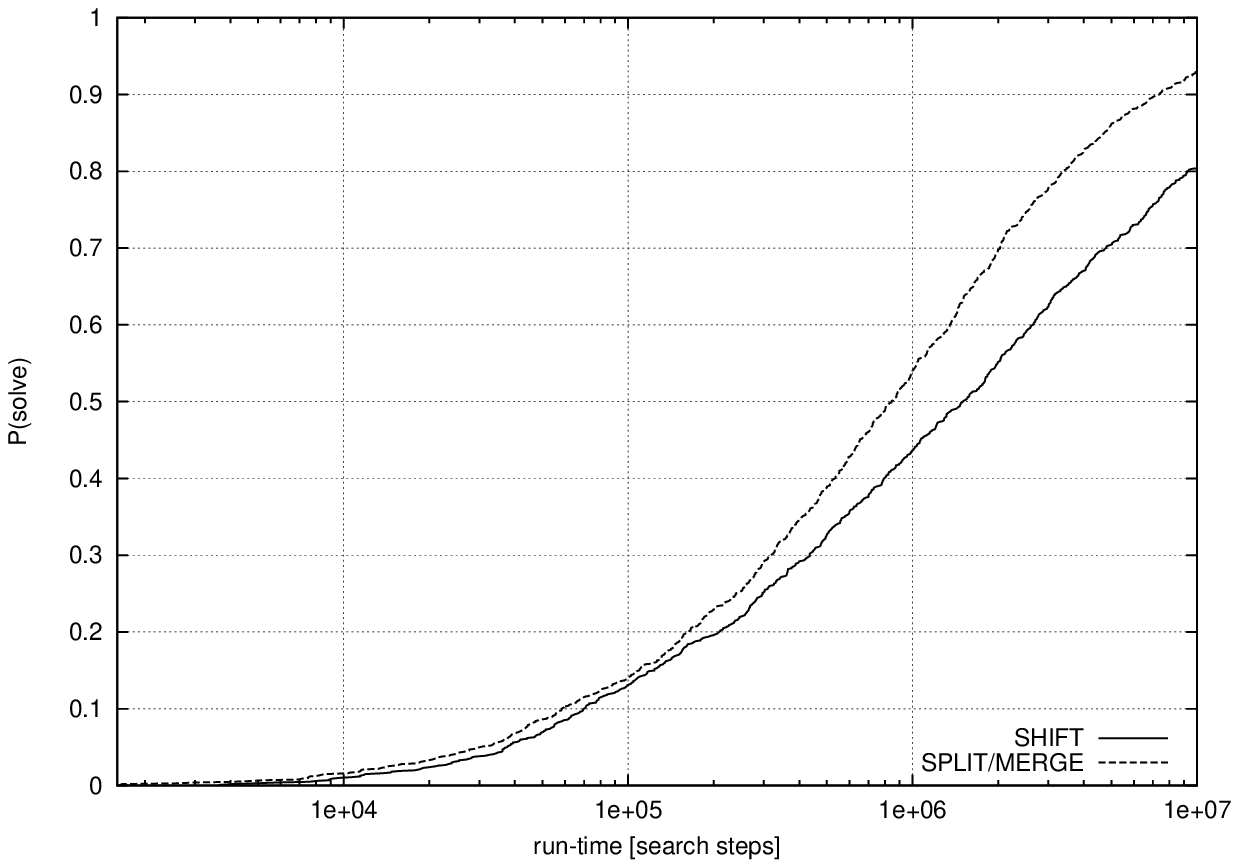}}
  \caption{Semi log-plot  of RLDs for  \texttt{GRASP}, with SHIFT  and SPLIT/MERGE neighbourhood operators, applied  to 100
    Uniform, Uniform  scaled, Normal, Normal  scaled, Normal distributed  CS instances, based  on 10
    runs per instance.} 
    \label{fig:grasp_prob}
\end{figure}

The first evaluation, whose results are reported in Figure~\ref{fig:grasp_prob} and
Table~\ref{tab:grasp_prob}, regards the behaviour of \texttt{GRASP} adopting the SPLIT/MERGE
($N_{s/m}$) or the SHIFT ($N_s$) neighbourhood relation in the local search phase. We set the number of agents to
15, the walk probability of the RII procedure to 0.7, and a cutoff run-length to $10^7$ operations. In particular, for each instance
of the problem we computed the solution quality obtained with \texttt{GRASP}, computed as the ratio between
the optimal solution value and the \texttt{GRASP} best solution value. The cutoff run-length limited the
number of operations of the \texttt{GRASP} algorithm. In particular, \texttt{GRASP} ends or when the solution quality
is 1 or when the number of computed operations is greater then the cutoff run-length. The operations
taken into account are the sum of the visited nodes during the construction and local search phase.
Hence, \texttt{maxiter} is set to $+ \infty$ and the \texttt{GRASP} stopping criteria is based on the solution
quality and number of operations.

We generated 100 problem instances for each distribution type (Uniform, Uniform scaled, Normal,
Normal scaled and Normally distributed) of the characteristic function. For
each instance 10 different runs of the \texttt{GRASP} algorithm were executed. Figure~\ref{fig:grasp_prob}
plots the graphs of the RLD for \texttt{GRASP} about each distribution of the characteristic function. Each
graph reports the curves corresponding to the cumulative empirical run-length distribution when the
local search uses the SPLIT/MERGE and SHIFT neighbourhood operators. As we can see  the SPLIT/MERGE
neighbourhood operator is more robust than the SHIFT operator and permits \texttt{GRASP} to find good
solutions more quickly.

An insight view of the experiment is reported in Table~\ref{tab:grasp_prob}, where some
descriptive statistics for the RDLs shown in Figure~\ref{fig:grasp_prob} are indicated. The first
column  reports the
adopted distribution for the characteristic function (Uniform (U), Uniform scaled (US), Normal (N),
Normal scaled (NS), and Normally distributed (ND)); the second column indicates the used
neighbourhood operator; \texttt{mean}, \texttt{min}, \texttt{max} and \texttt{stddev} indicate,
respectively, the mean, the minimum, the maximum and the  standard deviation of the operations
number over the 1000 runs (10 different \texttt{GRASP} execution for problem instance); \texttt{vc} denotes
the \emph{variational coefficient} ($vc = stddev / mean$); $q_{0.75}/q_{0.25}$ is a quantile
ratio; and \texttt{\#opt} is the number of runs \texttt{GRASP} found the optimal solution within the cutoff
run-length operations. From these statistics becomes more evident the improvement obtained adopting
the SPLIT/MERGE operator in the local search phase.

\begin{table}
  \centering
\begin{tabular}{|c|c|rrrrccc|}
  \hline
  Dist & Op &  \texttt{mean} \ & \texttt{min} & \texttt{max} & \texttt{stddev} & vc &
  $q_{0.75}/q_{0.25}$ & \texttt{\#opt}\\ \hline
 \multirow{2}{*}{U} & S & 20318.4 & 3698 & 279817 & 26048.4 & 1.28 & 3.46 & 1000\\
  & S/M & 48909.2 & 1658 & 1127280 & 89177.1 & 1.82 & 6.64 & 1000\\ \hline
  \multirow{2}{*}{US} & S & 8844716.9 & 33292 & 10006969 & 2565704.5 & 0.29 & 1.00 & 213\\
&  S/M & 4422351.4 & 5958 & 10138012 & 3719544.2 & 0.84 & 7.76 & 781\\ \hline
\multirow{2}{*}{N} & S & 6737.1 & 4438 & 15017 & 2134.0 & 0.32 & 1.76 & 1000\\
&  S/M & 3057.6 & 2280 & 6683 & 600.5 & 0.20 & 1.23 & 1000\\ \hline
\multirow{2}{*}{NS} & S &  8100474.3 & 6249 & 10006889 & 3299610.4 & 0.41 & 1.34 & 307\\
&  S/M & 3923760.0 & 2681 & 10076751 & 3496925.3 & 0.89 & 7.22 & 845\\ \hline
\multirow{2}{*}{ND} & S & 3461246.2 & 3637 & 10006795 & 3846373.4 & 1.11 & 22.78 & 803\\
 &  S/M & 2131138.6 & 1633 & 10004863 & 2903255.4 & 1.36 & 10.65 & 929\\ \hline
\end{tabular}
\caption{Descriptive statistics for the RLDs shown  in Figure~\ref{fig:grasp_prob}; $vc=stddev/mean$ denotes the
variation coefficient, and $q_{0.75}/q_{0.25}$ the quantile ratio, where  $q_x$ denotes the
$x$-quantile.}
\label{tab:grasp_prob}
\end{table}

\begin{figure}
  \centering
  \subfigure[Uniform]{\includegraphics[width = 0.49\textwidth]{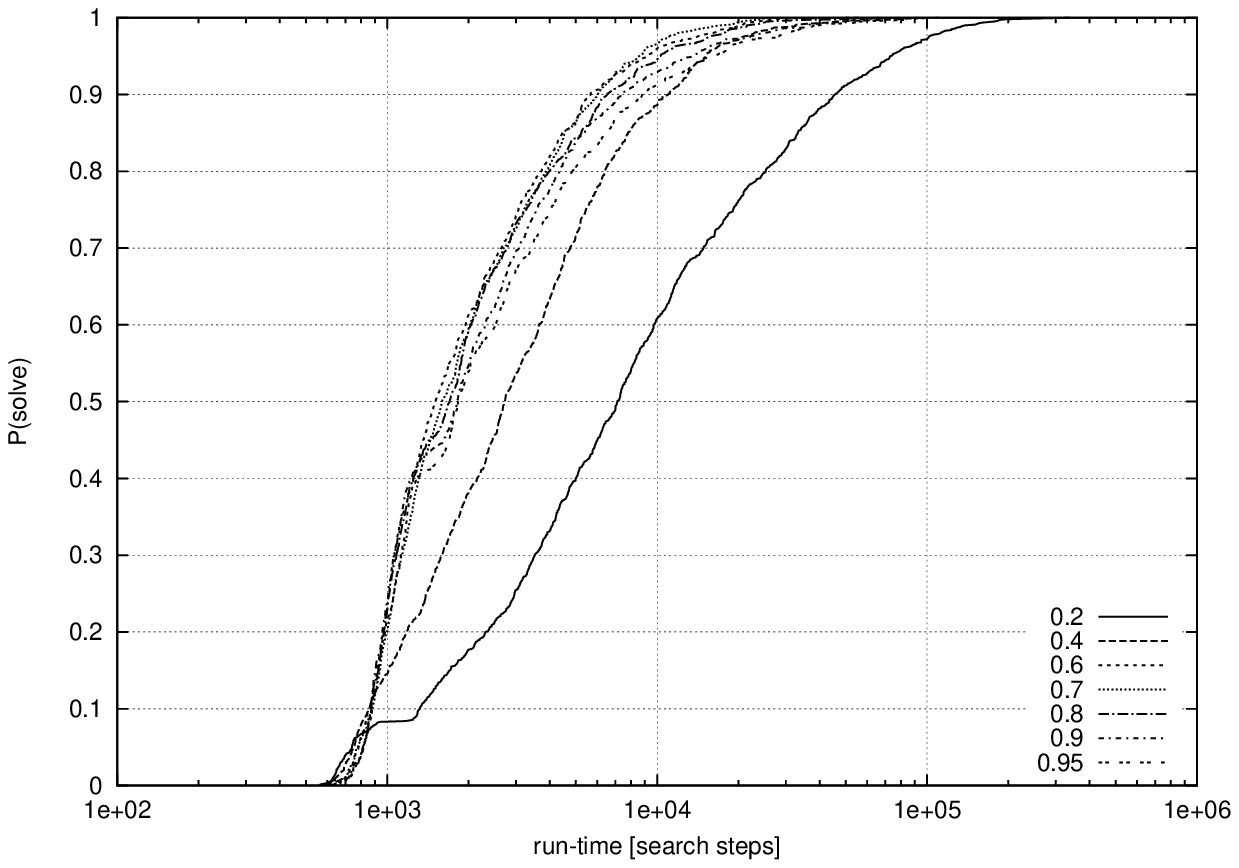}}
  \subfigure[Uniform scaled]{\includegraphics[width = 0.49\textwidth]{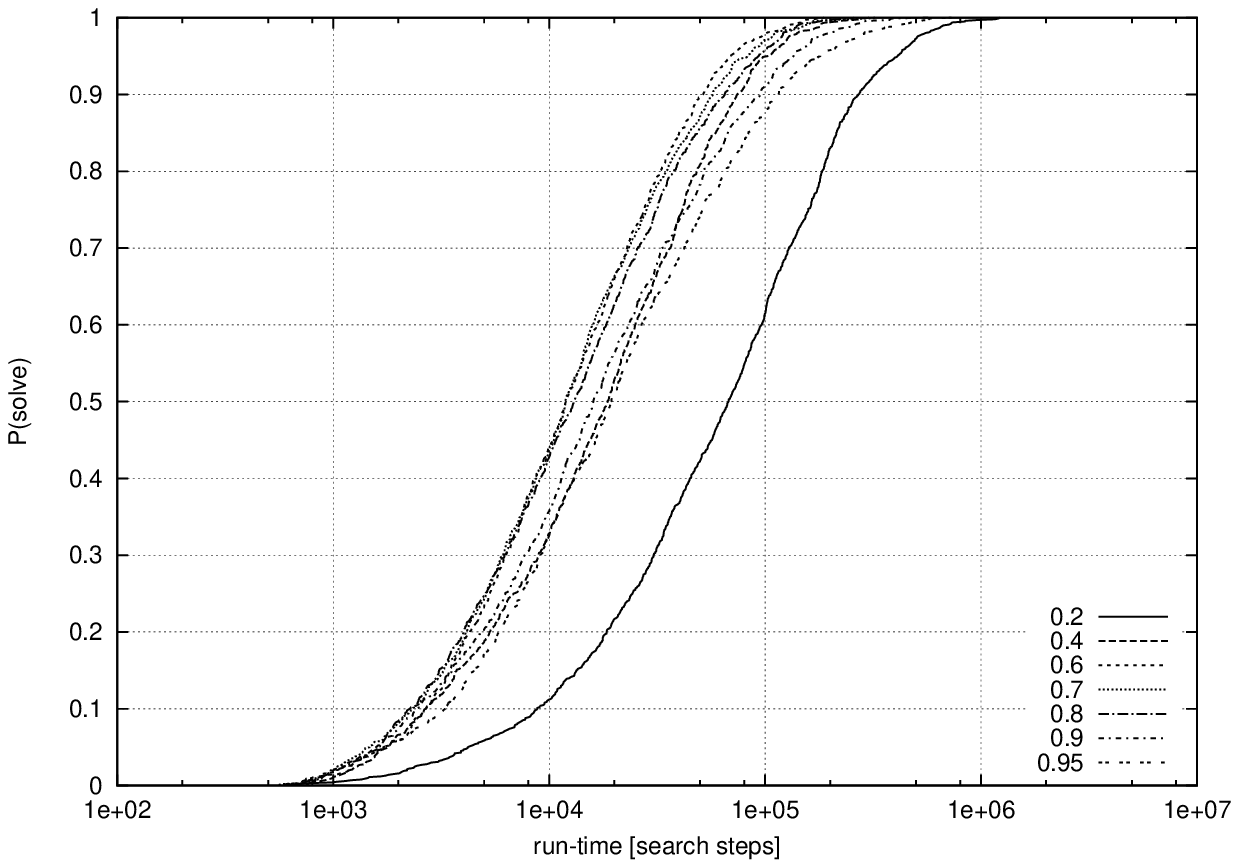}}
  \subfigure[Normal]{\includegraphics[width = 0.49\textwidth]{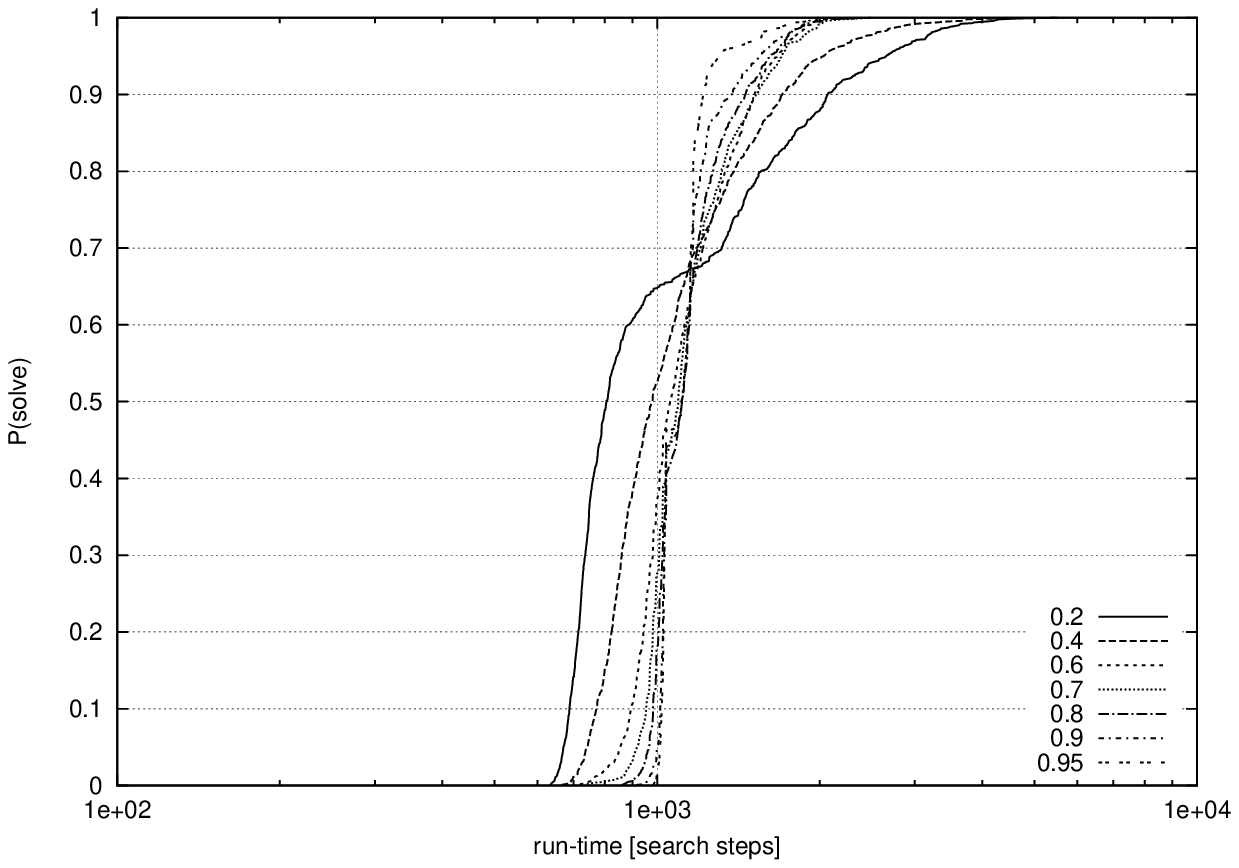}}
  \subfigure[Normal scaled]{\includegraphics[width = 0.49\textwidth]{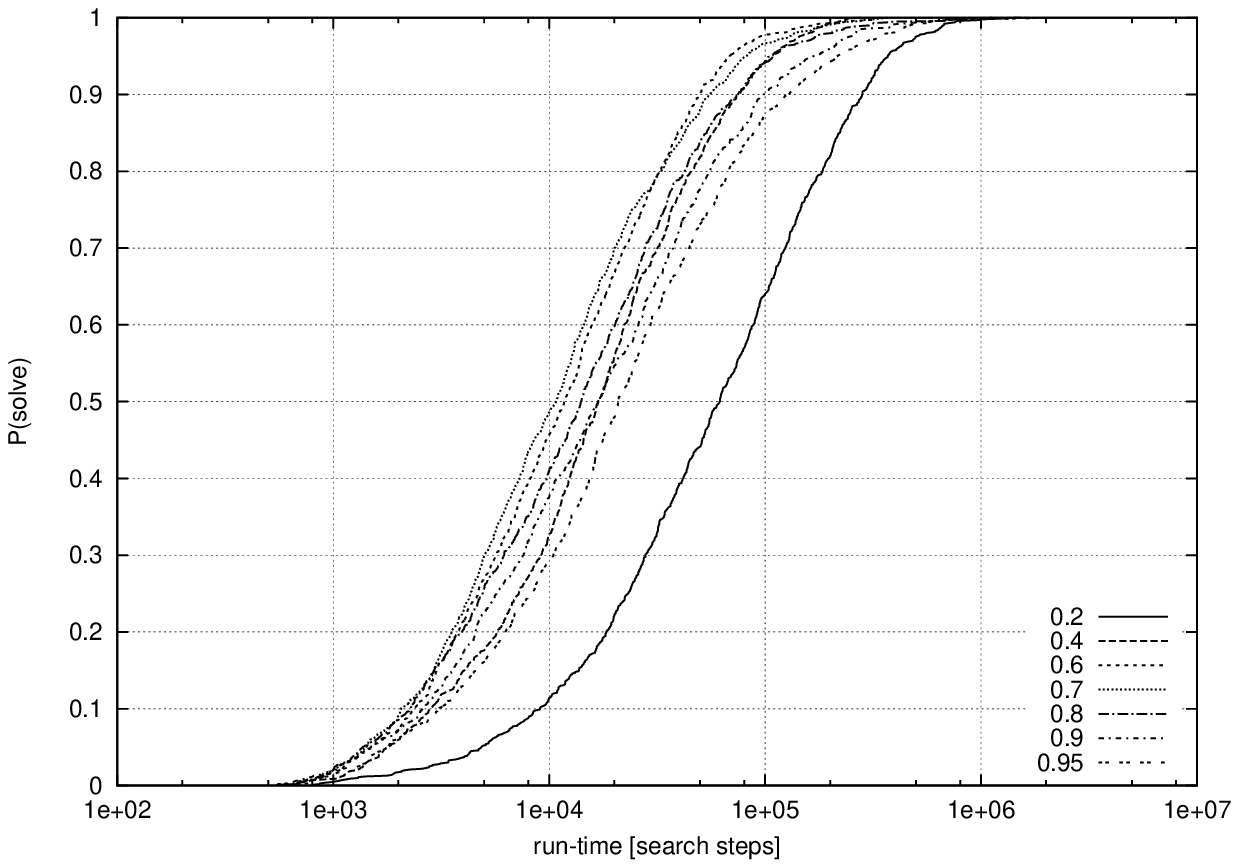}}
  \subfigure[Normal distributed]{\includegraphics[width = 0.49\textwidth]{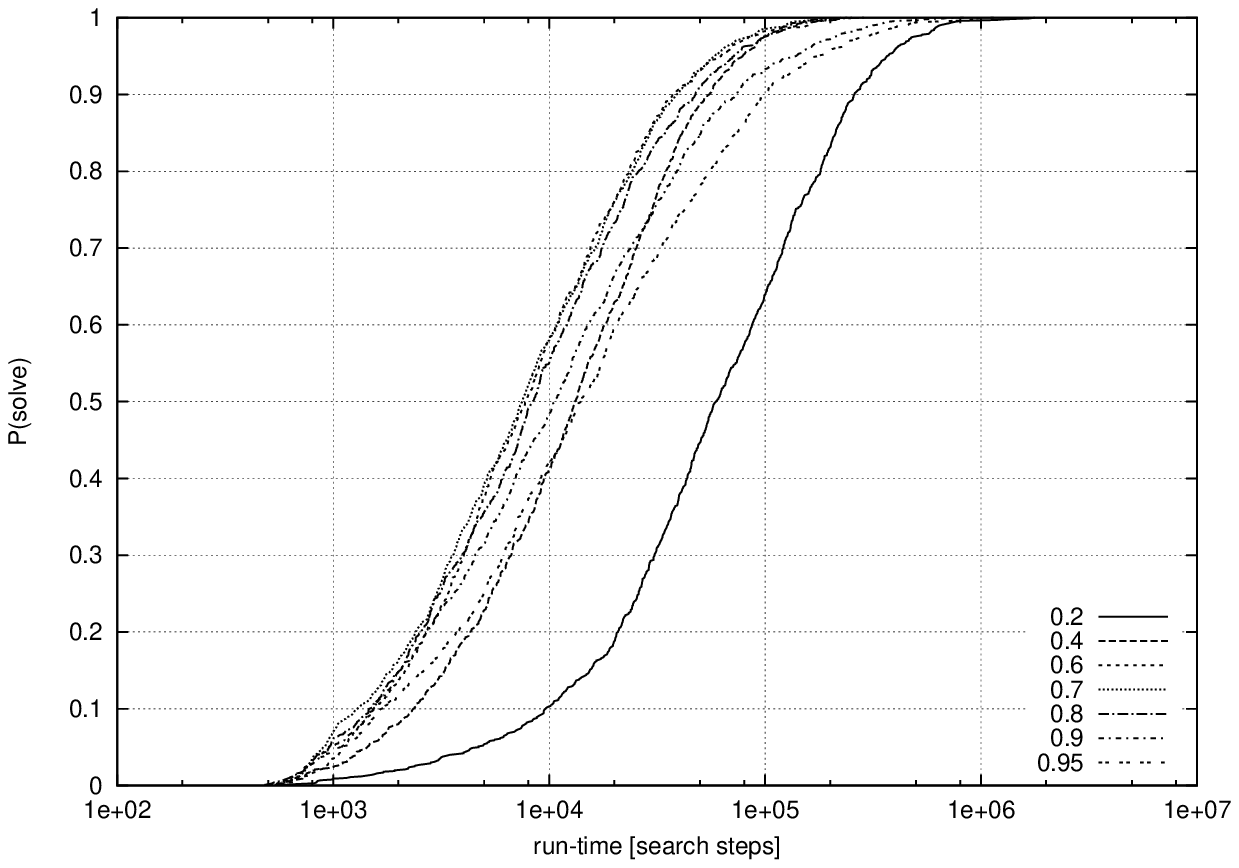}}
  \caption{Semi log-plot  of RLDs for  GRASP with SPLIT/MERGE neighbourhood operator, applied  to 100
    Uniform, Uniform  scaled, Normal, Normal  scaled, Normal distributed  CS instances, based  on 10
    runs per instance with different values of the walk probability for the RII procedure.} 
    \label{fig:grasp_wp}
\end{figure}

A second experiment was run in order to evaluate the impact of the walk probability value in the Randomised
Iterative Improvement procedure adopted as local search. The corresponding results are plotted in
Figure~\ref{fig:grasp_wp}. As in the previous experiment we run \texttt{GRASP} 10 times over 100 problem
instances for distribution and taking fixed the neighbourhood operator to SPLIT/MERGE. 
The best values are obtained with a walk probability equal to 0.6 or to
0.7. There is an increasing improvement for values ranging from 0.2 to 0.6/0.7; then the quality of the
found solutions starts to decrease for values ranging from 0.7 to 0.95. 

\section{GRASP with path-relinking for the CSG problem}
\label{sec:gpr}
Path-relinking is an intensification strategy, proposed in~\cite{Glover96tabusearch}, that explores 
trajectories  connecting  \emph{elite} solutions  obtained  by  tabu  search~\cite{Gl97} or  scatter
search~\cite{springerlink:10.1007/978-1-4419-1665-5_4}.

Given a set of elite solutions, paths among elite solutions in  the solution space are generated and traversed 
hoping to visit better solutions.  
Paths are generated taking into account the moves incorporating attributes of the guiding solution into the current one. 
Algorithm~\ref{alg:path-relinking}  reports the path-relinking  procedure
applied to a pair of solutions $x_s$ (starting solution) and $x_t$ (target solution), assuming that
$f(x_s) < f(x_t)$, where $f$ is the heuristic function computing the solution's value.

\begin{algorithm}[tb]
\caption{PATH-RELINKING}
\label{alg:path-relinking}
\begin{algorithmic}[1]
\REQUIRE{$x_s$ and $x_t$: starting and target solution such that $f(x_s) < f(x_t)$}
\ENSURE{best solution $x^*$ in path from $x_s$ to $x_t$}
\STATE $x = x_s$
\STATE $\mathcal D = \Delta(x,x_t)$
\WHILE{$\mathcal D \neq \emptyset$}
  \STATE select $m^*  \in \mathcal D$ such  that $f(x \oplus m^* )  < f(x \oplus m_i )$  for all $m^*
   \neq m_i  \in \mathcal D$
  \STATE $\mathcal D = \mathcal D \setminus \{m^*\}$
  \STATE $x = x \oplus m^*$
  \IF{$f(x) < f^*$} 
    \STATE $f^* = f(x)$
    \STATE $x^* = x$
  \ENDIF
\ENDWHILE
\STATE \textbf{return} $x^*$
\end{algorithmic}
\end{algorithm}

The algorithm  iteratively computes the symmetric difference $\Delta(x,x_t)$ between the current solution $x$ and the target one 
$x_t$ corresponding to the set of moves needed to reach $x_t$ from $x$.
At each step, the algorithm considers all the possible  moves $m \in \Delta(x, x_t)$ and selects the one
which result is the least cost solution, i.e. the one which minimises $f(x \oplus m)$, where $x \oplus m$ is the solution
resulting from applying move $m$ to solution $x$. The best move $m^*$ is made, producing solution $x \oplus m^*$. 
The algorithm terminates when $x_t$ is reached, i.e. when $\Delta(x,x_t)=\emptyset$, and returns the best solution $x^*$ obtained among the iterations.

Given two elite solutions $a$ and $b$, some of the alternatives to relink $a$ and
$b$~\cite{Resende05,springerlink:10.1007/978-1-4419-1665-5_4}  considered in this paper are:
\begin{itemize}
\item \emph{forward relink}: using $x_s = \min_{a,b} \{ f(a), f(b) \}$ and $x_t = \max_{a,b} \{ f(a), f(b) \}$;
\item \emph{backward  relinking}:  adopting $x_s = \max_{a,b} \{ f(a), f(b) \}$ and $x_t = \min_{a,b} \{ f(a), f(b) \}$;
\item \emph{back  and forward relinking}: both  different forward and backward trajectories are explored.
\end{itemize}

Path-relinking represents a  major enhancement  to  the  basic  GRASP  procedure, leading  to  significant
improvements in solution time and quality, firstly proposed in~\cite{Laguna99graspand}. 
The path-relinking intensification strategy adopted in this paper is applied to each local optimum obtained after the local search phase.

\begin{algorithm}[tb]
\caption{GRASP+PR}
\label{alg:graspwpr}
\begin{algorithmic}[1]
\REQUIRE{$v$: the characteristic function;\\ $A$: the set of $n$ agents;\\
\texttt{maxIter}: maximum number of iterations; \\
\texttt{neighOP}: neighbourhood operator; \\
\texttt{maxElite}: max pool dimension; \\
\texttt{riiSteps}: max non improving search steps for the RII procedure; \\
\texttt{wp}: RII walk probability}
\ENSURE{solution $\widehat{C} \in \mathcal M(A)$}
\STATE iter $= 0$
\STATE $P = \emptyset$
\WHILE{iter $<$ \texttt{maxiter}}
\STATE $C = $ \texttt{GreedyRandomisedConstruction()} \emph{/* lines 4-13 of the
Alg.~\ref{alg:grasp} */}
  \STATE $C = $ \texttt{RandomisedIterativeImprovement(}$C$, \texttt{wp}, \texttt{riiSteps},
  \texttt{neighOP})
  \IF{iter $\geq 1$}
  \FORALL{ $x \in P$}
		\STATE determine which ($C$ or $x$) is the initial and which is the target
		\STATE $x_p$ = \texttt{PathRelinking}($x_s$, $x_t$)
		\STATE update the elite set with $x_p$ 
		\IF{$v(x_p) > v(\widehat{C})$}
			\STATE $\widehat{C} = x_p$
		\ENDIF
	\ENDFOR
      \STATE update the elite set $P$ with $C$
  \ELSE
    \STATE insert $C$ into the elite set $P$
  \ENDIF
  \STATE iter = iter + 1
\ENDWHILE
\STATE \textbf{return} $\widehat{C}$
\end{algorithmic}
\end{algorithm}

The algorithm adopts a pool of \texttt{maxElite} elite solutions that is originally empty. Then,
each locally optimal solution obtained by a local search is considered as a candidate to be inserted
into the pool if it is different from every solution currently contained into the pool. The strategy
adopted in this paper is the following. If the pool already has \texttt{maxElite} solutions, the
candidate is inserted into the pool if it is better than the worst of them, that is then removed from
the pool. If the pool is not full, the candidate is simply inserted.

Algorithm~\ref{alg:graspwpr} reports GRASP with path-relinking for the CSG problem, referred in the
following as \texttt{GRASP+PR}, where a new step
to the construction and local search phase is inserted. The path-relinking algorithm is applied to
the solution returned by local search and to all the solutions from the pool. Improving solutions
along the tajectories are considered as candidates for insertion into the pool.

\subsection{GRASP with path-relinking evaluation}

\begin{figure}
  \centering
  \subfigure[Uniform]{\includegraphics[width = 0.49\textwidth]{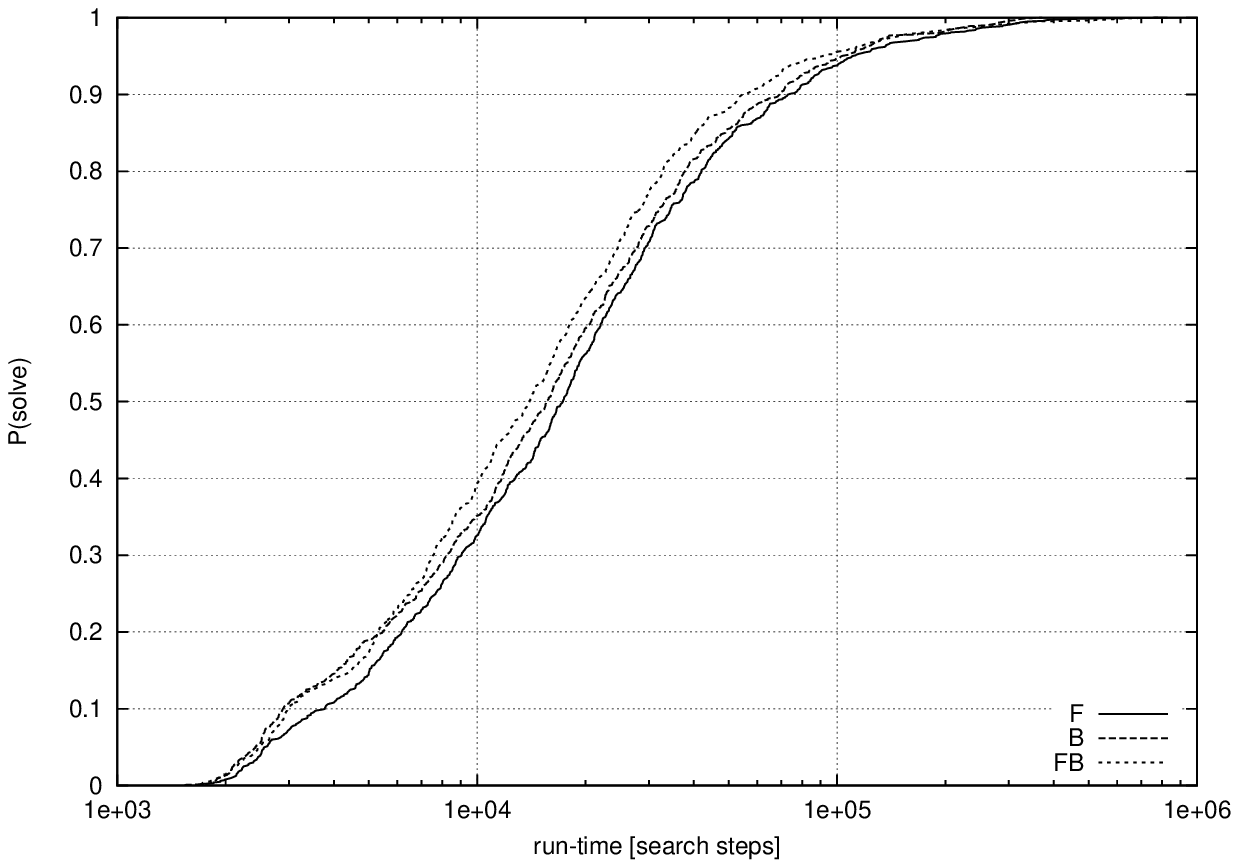}}
  \subfigure[Uniform scaled]{\includegraphics[width = 0.49\textwidth]{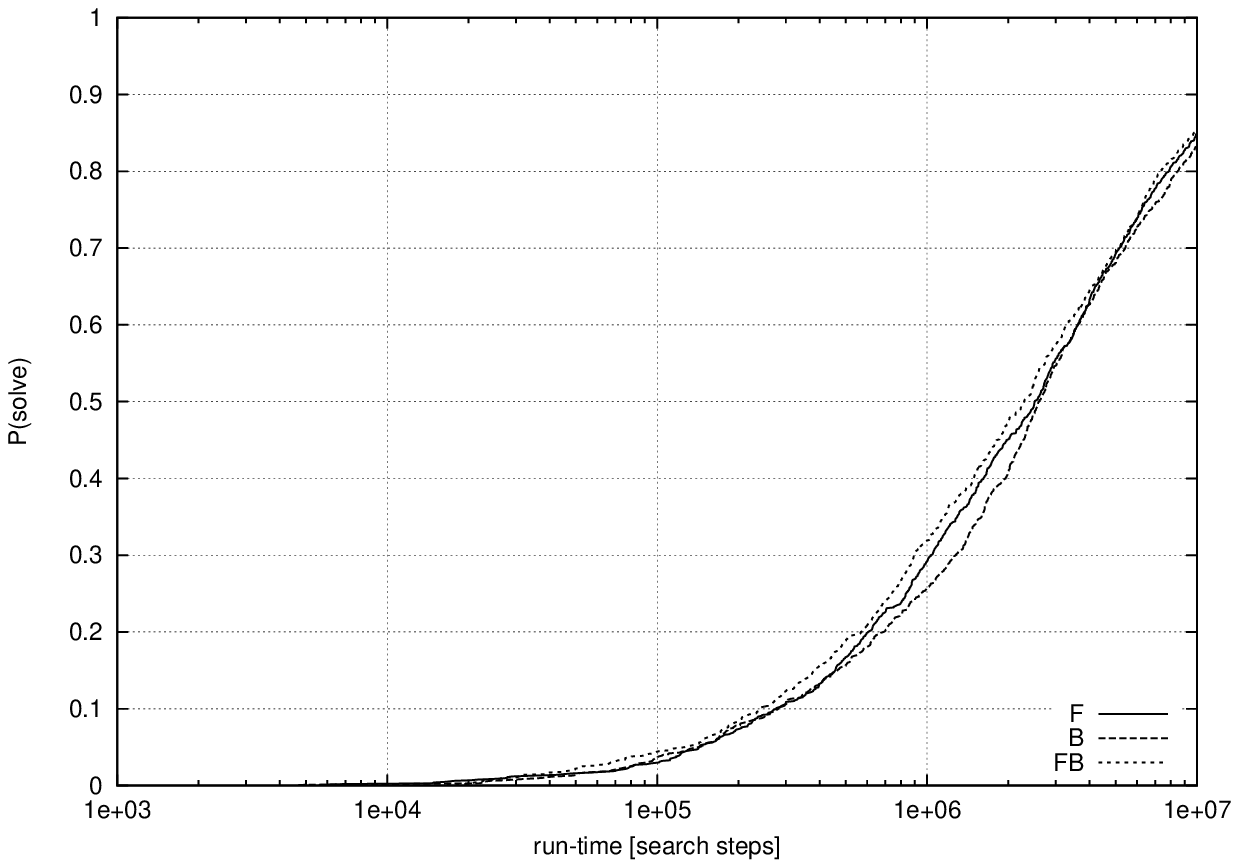}}
  \subfigure[Normal]{\includegraphics[width = 0.49\textwidth]{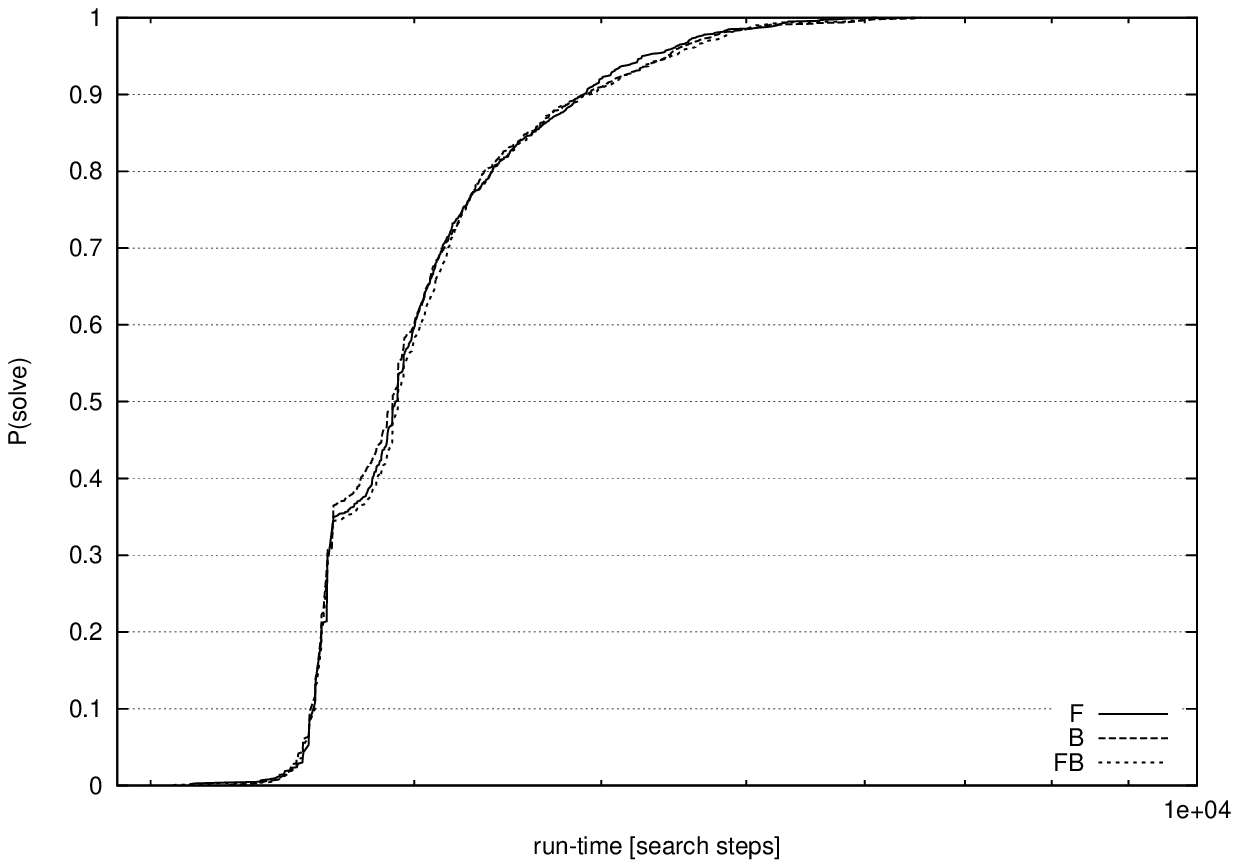}}
  \subfigure[Normal scaled]{\includegraphics[width = 0.49\textwidth]{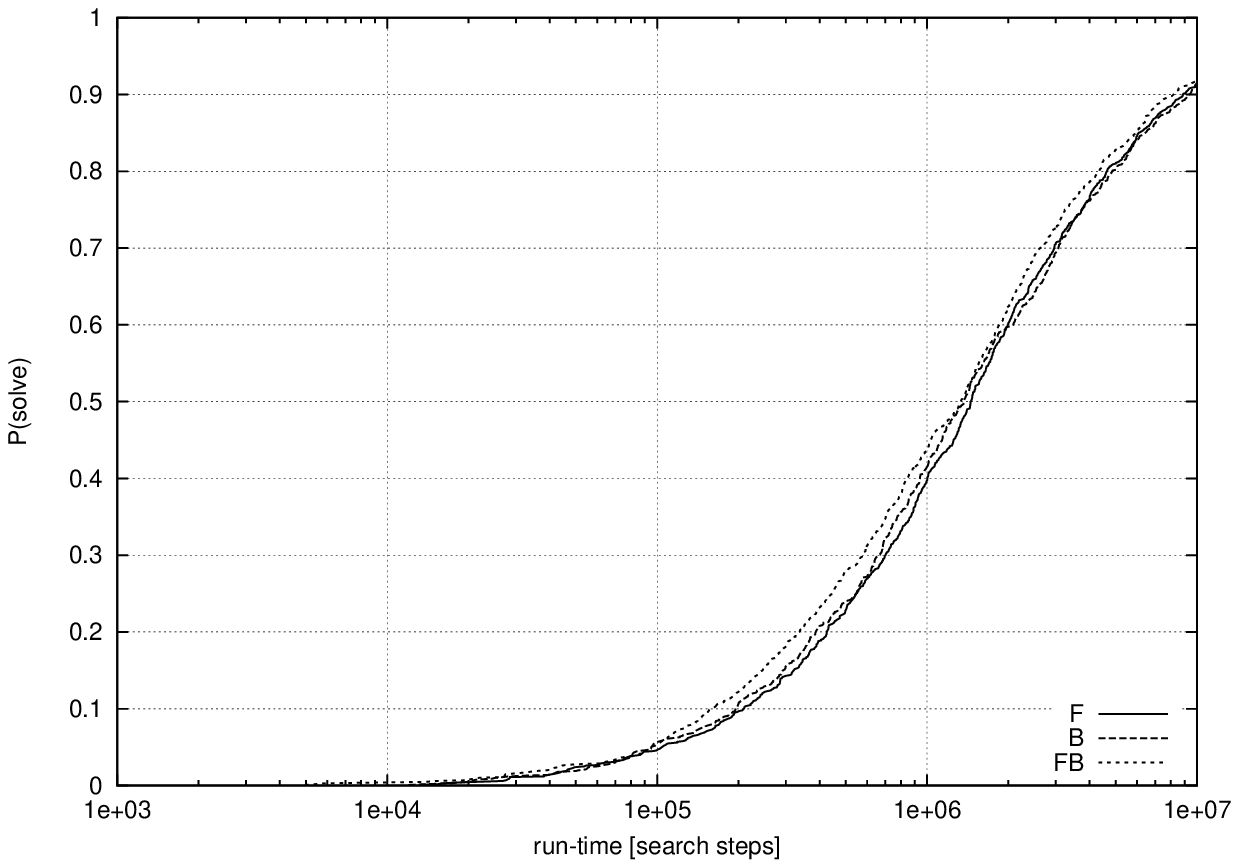}}
  \subfigure[Normal distributed]{\includegraphics[width = 0.49\textwidth]{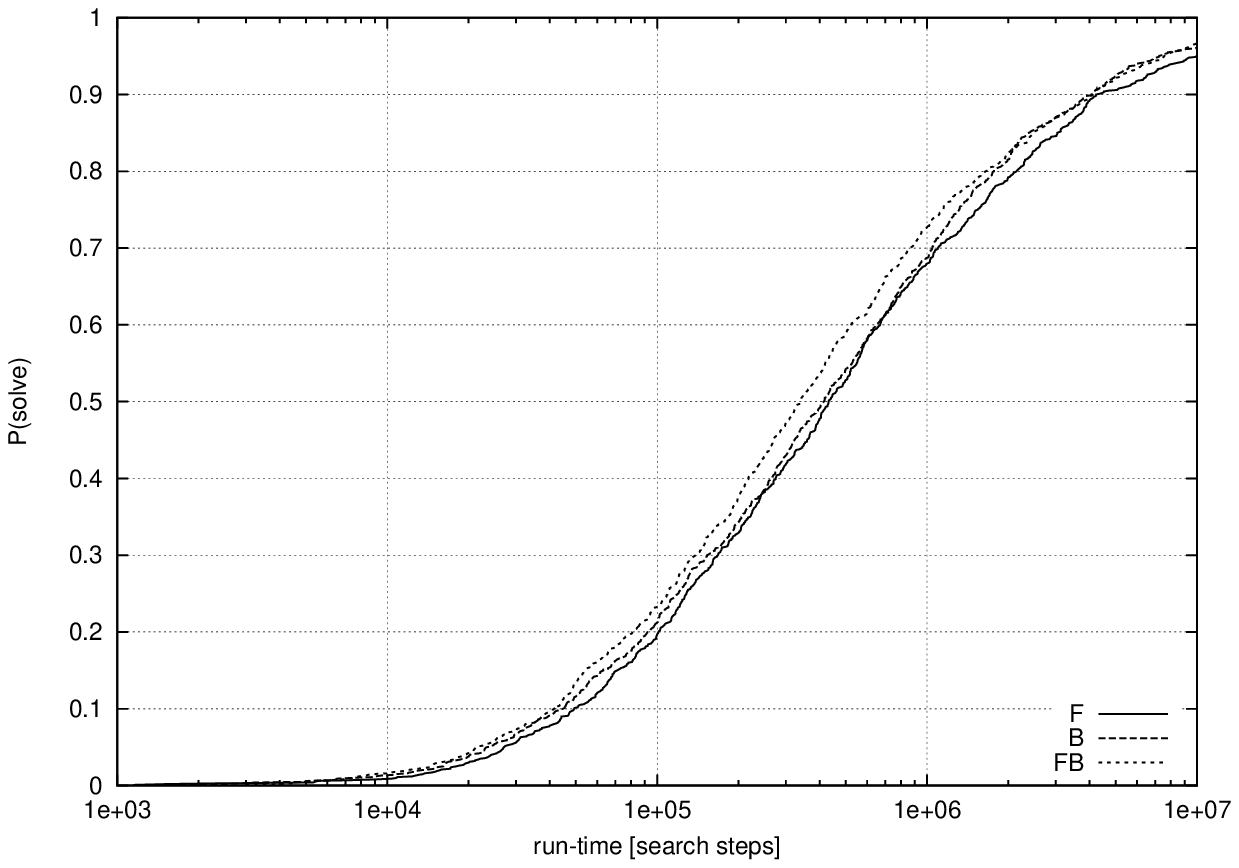}}
  \caption{Semi log-plot of RLDs for \texttt{GRASP-PR},  with SPLIT/MERGE neighbourhood operator and FORWARD
  (F), BACKWARD (B), and
    FORWARD-BACKWARD (FB) relinking strategy, applied to 100 Uniform, Uniform scaled, Normal, Normal scaled, Normal
    distributed CS instances, based on 10 runs per instance.}
    \label{fig:gpr_type}
\end{figure}

The first part of the evaluation of \texttt{GRASP+PR} for CSG regarded its efficacy by varying the relinking strategy and the
size of the pool of elite solutions. 
In the first experiment we investigated the \texttt{GRASP+PR} efficacy by varying the relinking
strategy. Figure~\ref{fig:gpr_type} plots the graphs corresponding to the
RLDs for \texttt{GRASP+PR} adopting the SPLIT/MERGE neighbourhood operator and different relinking
strategies. The setup of the experiment is the same as that used in the previous \texttt{GRASP}
evaluations, reported in Section~\ref{sec:grasp_eval}:  100
problem instances for each distribution and 10 \texttt{GRASP+PR} runs per instance; the number of
agents was set to 15, the walk probability of RII to 0.7,  and the cutoff run-length to $10^7$ operations. For each instance the solution
quality obtained with \texttt{GRASP+PR} has been computed as the ratio between the optimal solution
value and the \texttt{GRASP+PR} best solution value. As we can see from the graphs reported in
Figure~\ref{fig:gpr_type} the FORWARD/BACKWARD strategy is the most robust for all the distributions, while the FORWARD strategy 
seems to be the less beneficial one.
Table~\ref{tab:gpr_st} reports a descriptive statistics for the search space nodes visited
by \texttt{GRASP+PR} with a FORWARD relinking strategy, where \texttt{\#nodes} is the total number
of visited nodes; \texttt{\#construction}, \texttt{\#local} and \texttt{relinking} are,
respectively, the number of nodes visited in the construction, local search and relinking phase;
\texttt{iter} represents the mean value of the iterations required by the \texttt{GRASP+PR} algorithm
to end a single run.

\begin{table}
\begin{tabular}{|c|rrrrr|} 
\hline
Dis. & \texttt{\#nodes} & \texttt{\#construction} & \texttt{\#local} & \texttt{\#relink} &
\texttt{iter} \\ \hline
U & 33109.5 & 6745.0 (20.4\%) & 25578.4 (77.2\%) & 786.0 (2.4\%) & 13.2 \\
US & 3789510.5 & 363175.4 (9.6\%) & 3334008.3 (88.0\%) & 92326.8 (2.4\%) & 893.3\\
N & 3061.9 & 605.2 (19.8\%) & 2456.7 (80.2\%) & 0 (0.0\%) & 1\\
NS & 2732970.0  & 350673.3 (12.8\%) & 2268367.5 (83.0\%) & 113929.2 (4.17\%) & 853.0\\
ND & 1528595.9 &  301562.6 (19.7\%) & 1105793.2 (72.3\%) & 121240.1 (7.9\%) & 699.0\\
 \hline
\end{tabular}
\caption{Descriptive statistics for the search space  nodes visited by \texttt{GRASP+PR} with FORWARD
relinking strategy and SPLIT/MERGE neighbourhood operator.}
  \label{tab:gpr_st}
\end{table}

\begin{figure}
  \centering
  \subfigure[Uniform]{\includegraphics[width = 0.49\textwidth]{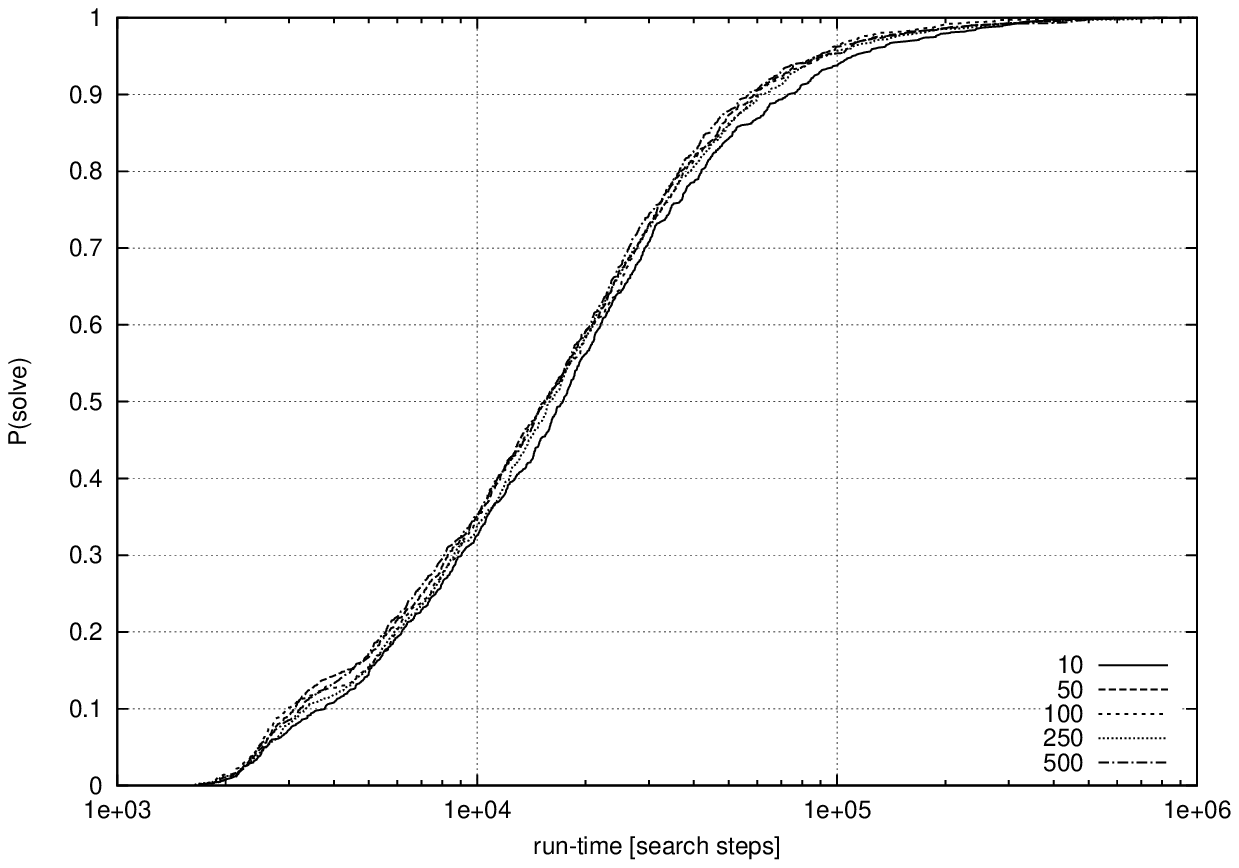}}
  \subfigure[Uniform scaled]{\includegraphics[width = 0.49\textwidth]{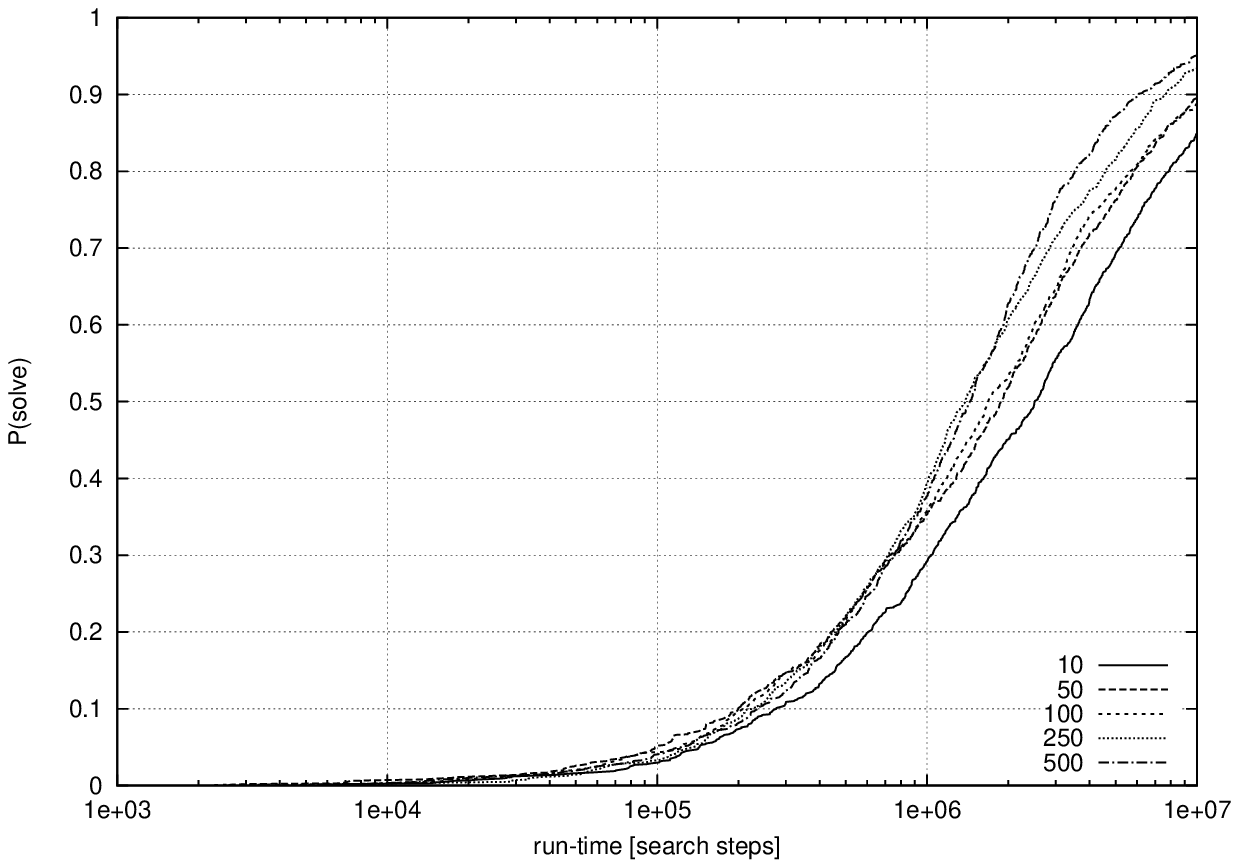}}
  \subfigure[Normal]{\includegraphics[width = 0.49\textwidth]{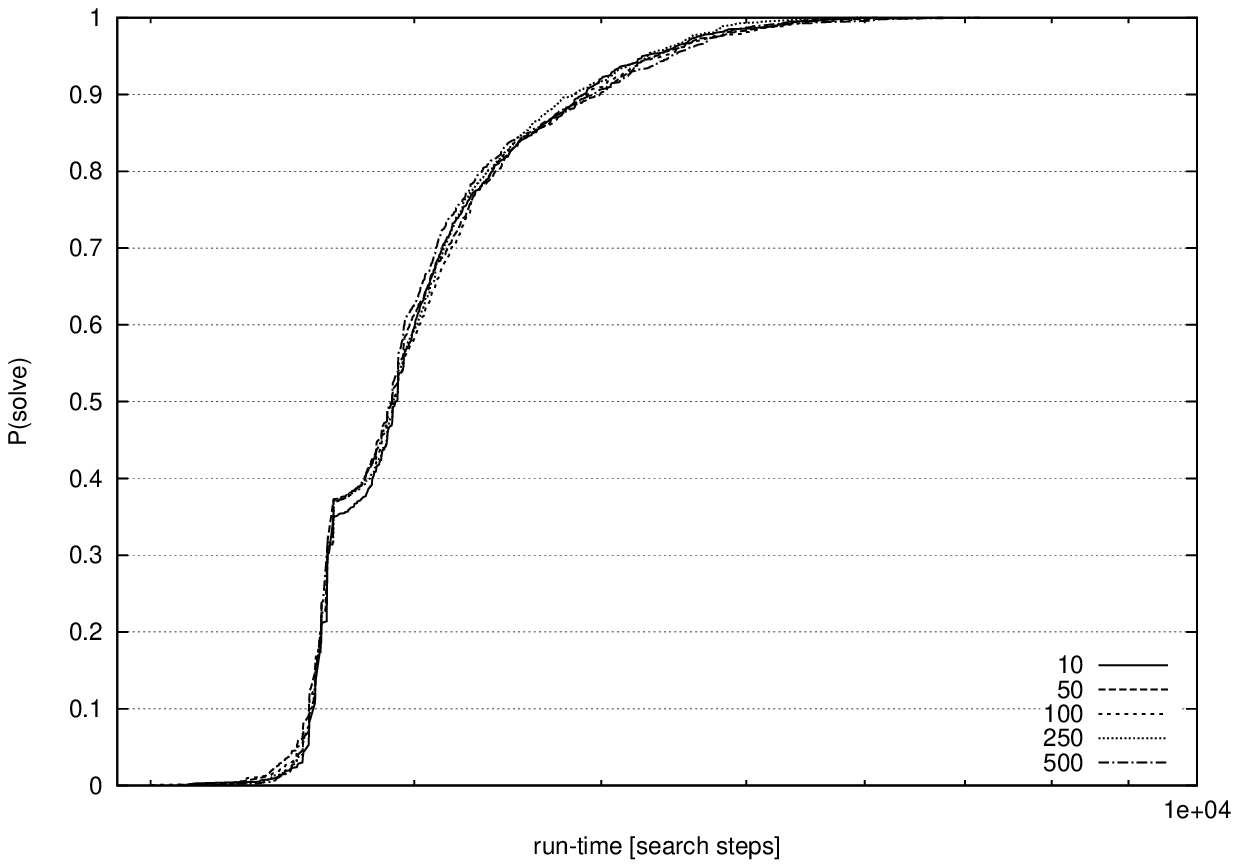}}
  \subfigure[Normal scaled]{\includegraphics[width = 0.49\textwidth]{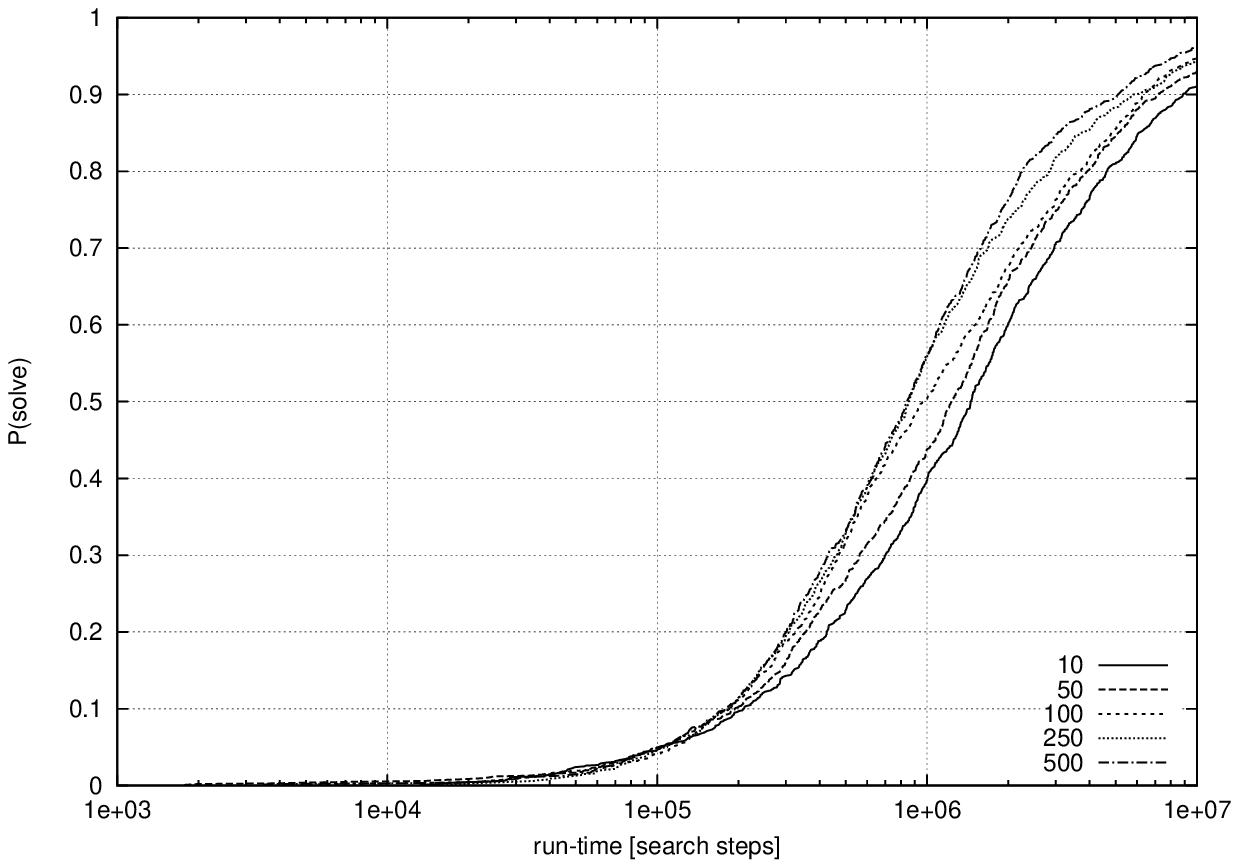}}
  \subfigure[Normal distributed]{\includegraphics[width = 0.49\textwidth]{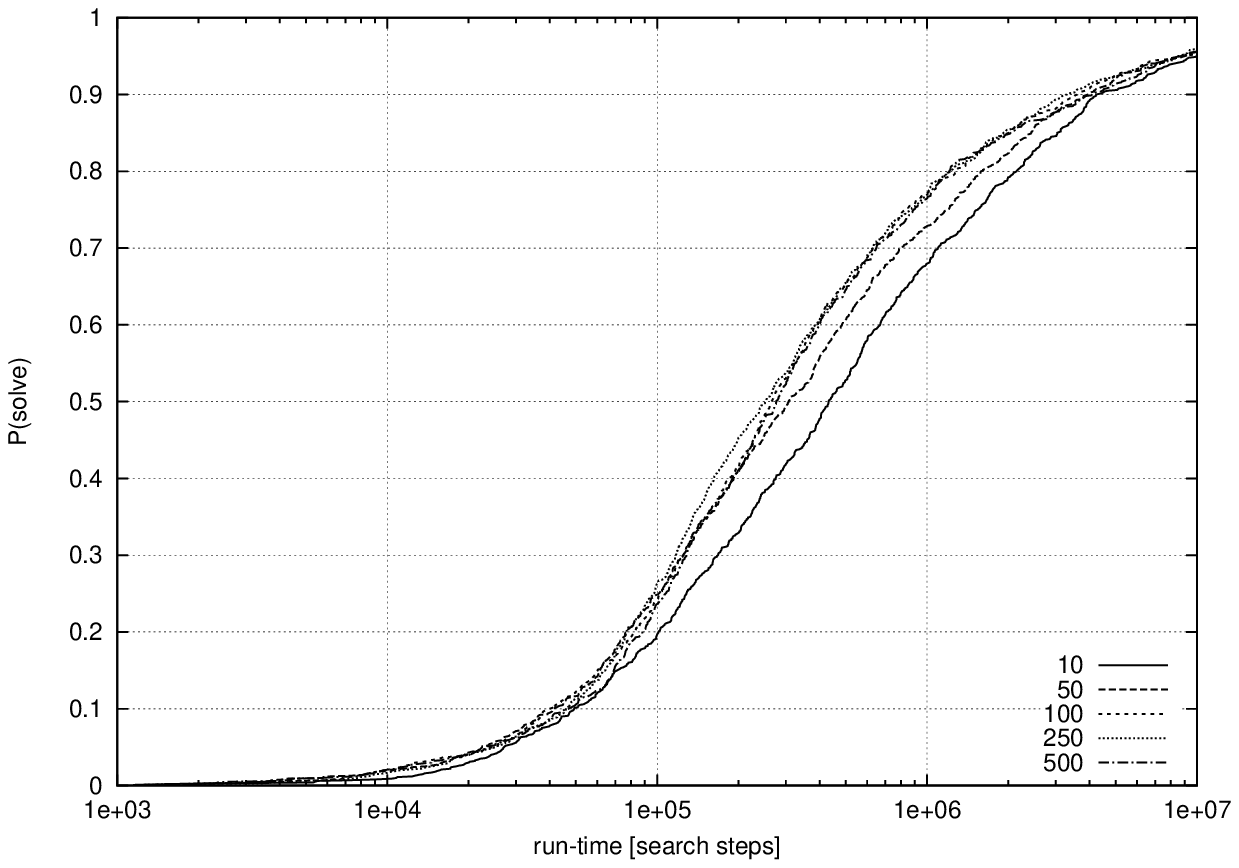}}
  \caption{Semi log-plot of RLDs for \texttt{GRASP+PR}  with SPLIT/MERGE neighbourhood operator and FORWARD
  relinking strategy, applied to 100 Uniform, Uniform scaled, Normal, Normal scaled, Normal
    distributed CS instances, based on 10 runs per instance. Curves refer to different values for the
    size of the pool of elite solutions.} 
    \label{fig:gpr_pool}
\end{figure}

In the second experiment of \texttt{GRASP+PR} we evaluated its efficacy by varying the size of the pool of elite solutions. Adopting the same
setting of the last experiment, we fixed the FORWARD relinking strategy and let to range the
\texttt{maxElite} parameter from the values belonging to the set \{10, 50, 100, 250, 500\}.
Figure~\ref{fig:gpr_pool} plots the obtained RLDs showing that, for each distribution, the adoption of a large \texttt{maxiter} value let \texttt{GRASP+PR} to quickly find the best solution.

\begin{figure}
  \centering
  \subfigure[Uniform]{\includegraphics[width = 0.49\textwidth]{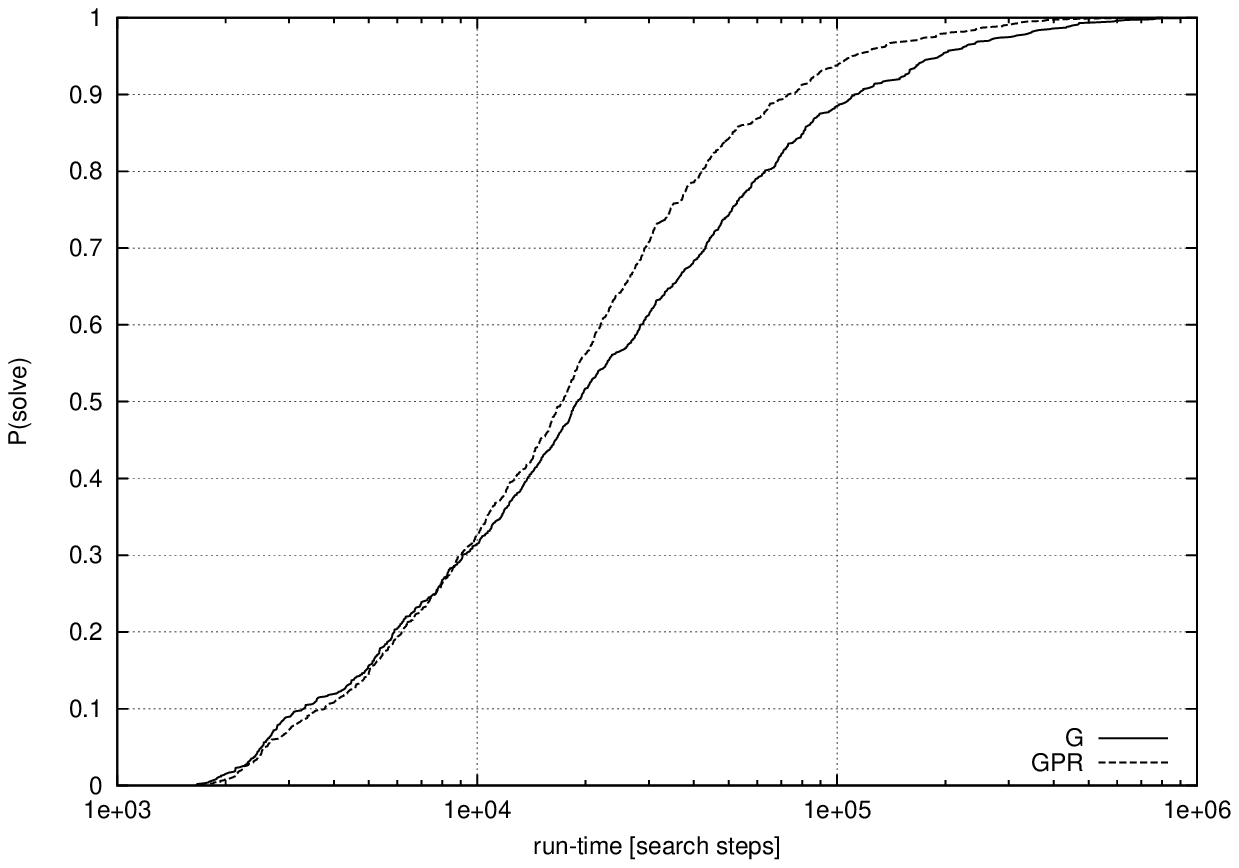}}
  \subfigure[Uniform scaled]{\includegraphics[width = 0.49\textwidth]{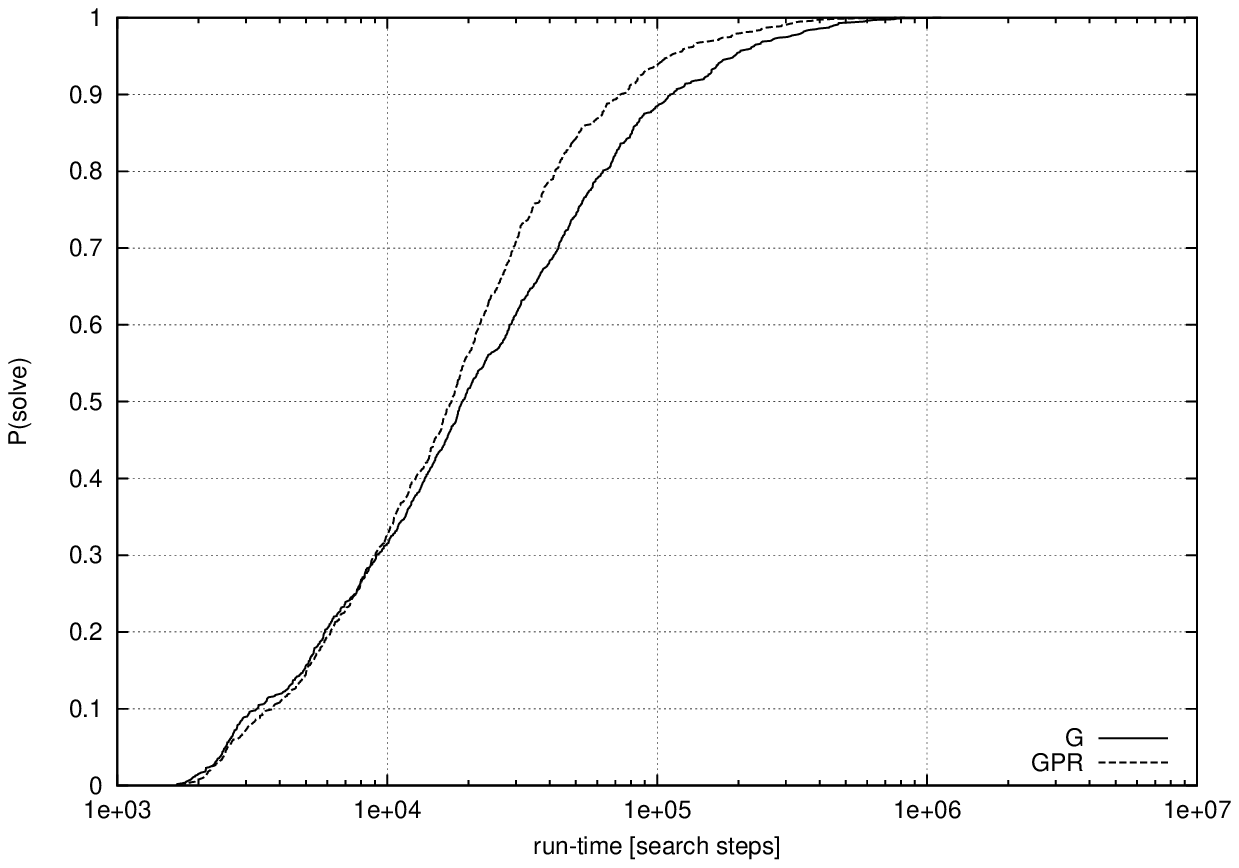}}
  \subfigure[Normal]{\includegraphics[width = 0.49\textwidth]{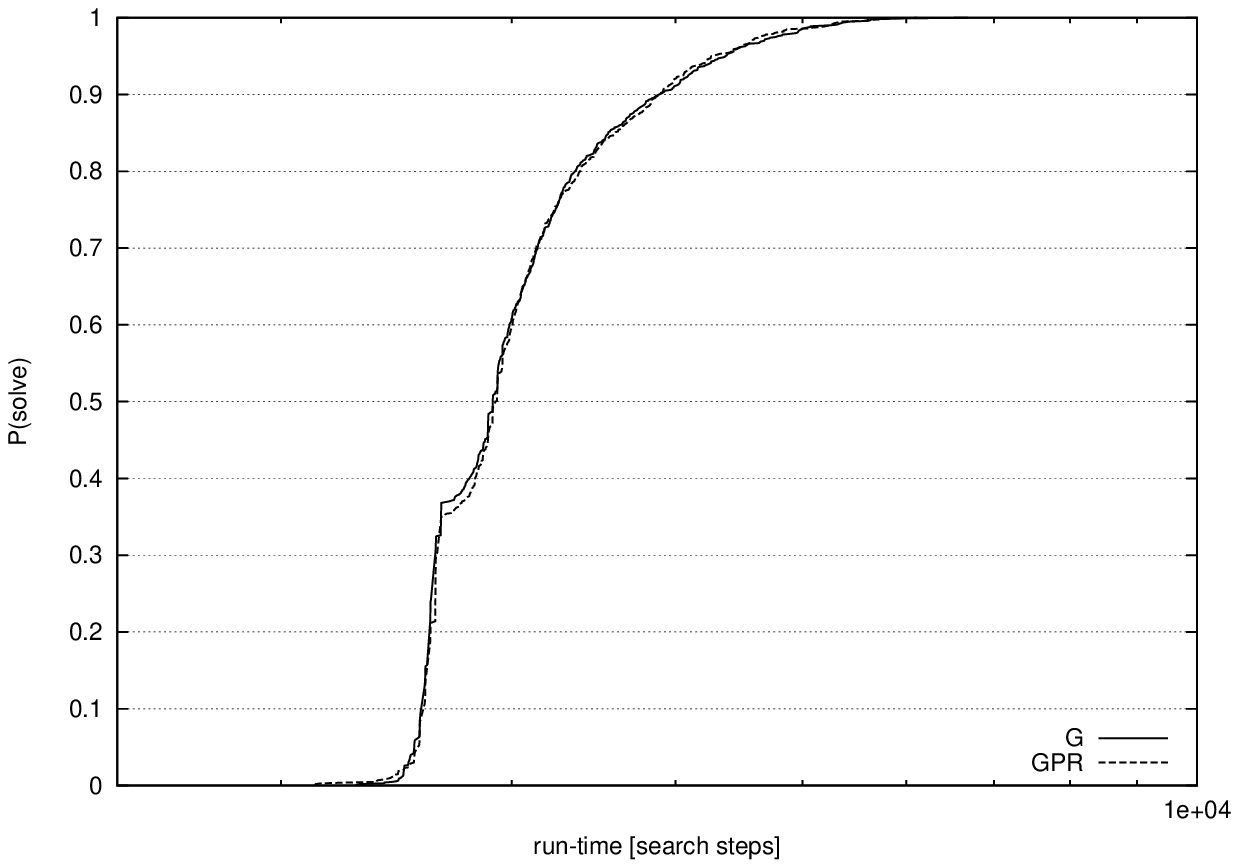}}
  \subfigure[Normal scaled]{\includegraphics[width = 0.49\textwidth]{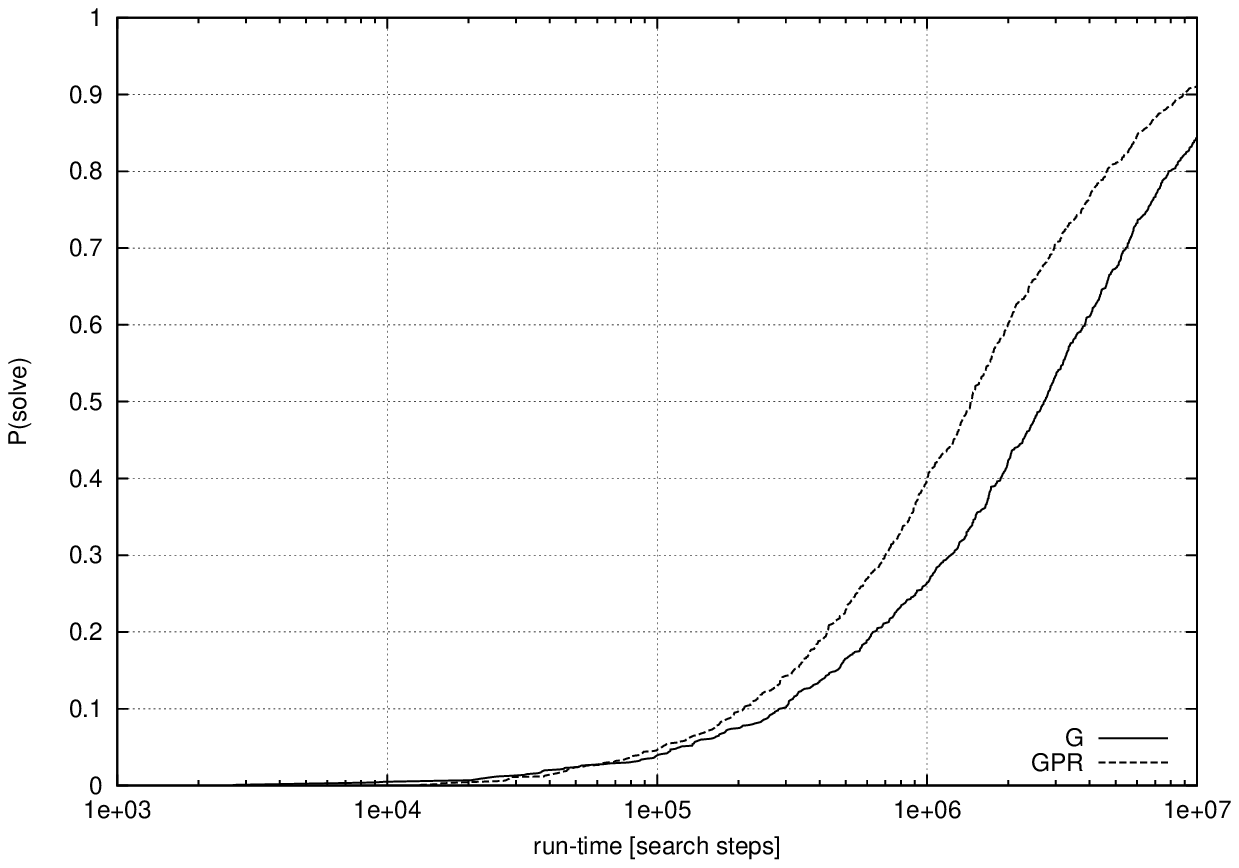}}
  \subfigure[Normal distributed]{\includegraphics[width = 0.49\textwidth]{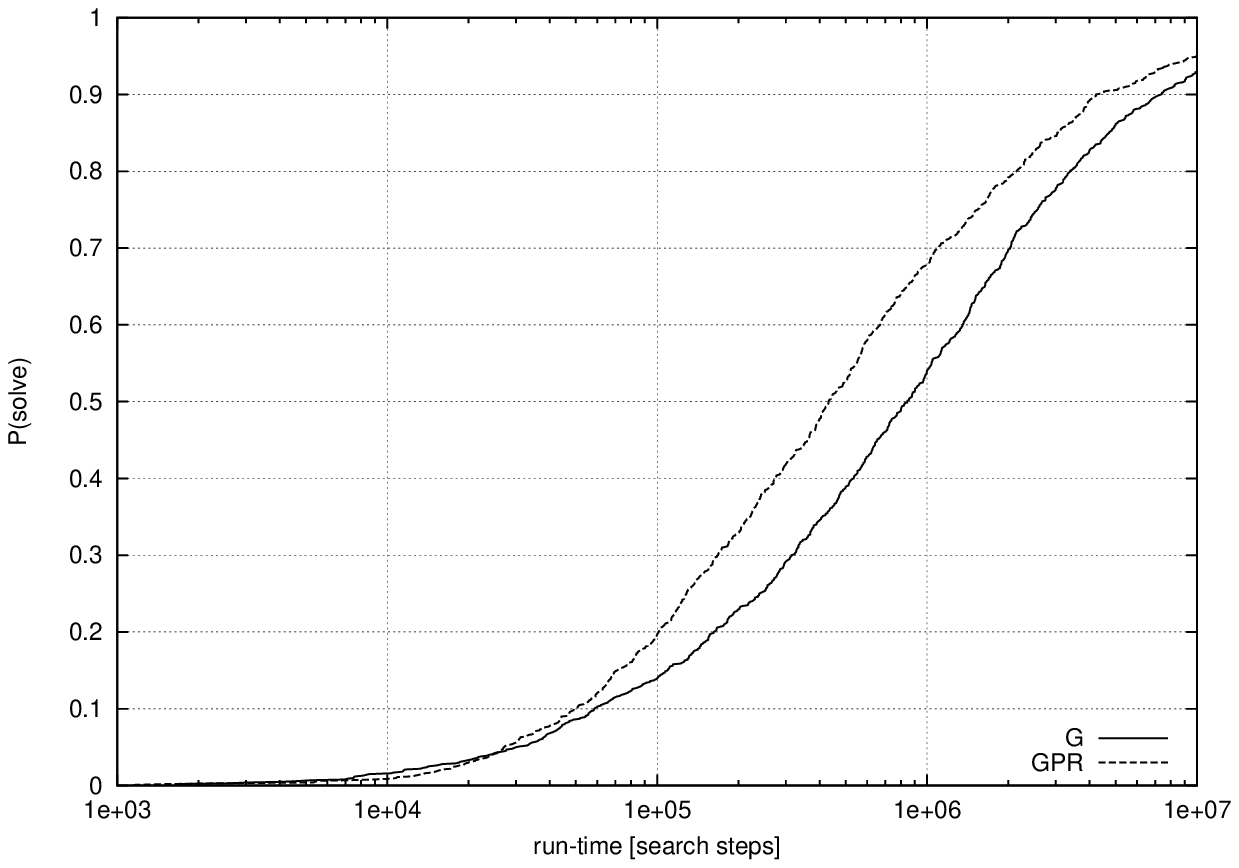}}
  \caption{Semi log-plot  of RLDs for GRASP and  GRASP+PR with FORWARD relinking strategy and SPLIT/MERGE
    neighbourhood operator, applied to  100 Uniform, Uniform scaled, Normal, Normal  scaled, Normal distributed CS
    instances, based on 10 runs per instance.} 
    \label{fig:ggpr_p}
  \end{figure}

The third experiment compared the behaviour of \texttt{GRASP} with respect to that of
\texttt{GRASP+PR} by considering solution qualities and the runtime performances. The setting of the
experiment is the same:  100 problem instances for each distribution and 10 \texttt{GRASP+PR} runs per instance; the number of
agents was set to 15, the walk probability of RII to 0.7,  the cutoff run-length to $10^7$
operations, the relinking strategy was FORWARD, the neighbourhood operator was SPLIT/MERGE, and the
pool size of the elite solutions to 10. Figure~\ref{fig:ggpr_p} plots the
obtained RDLs and the Table~\ref{tab:ggpr_p} reports the corresponding descriptive statistics.

\begin{table}
\begin{tabular}{|c|l|ccccccc|} 
\hline
Dist. & Alg. & \texttt{mean} & \texttt{min} & \texttt{max} & \texttt{stddev} & vc & $q_{0.75}/q_{0.25}$ & \texttt{\#opt}\\ \hline
\multirow{2}{*}{U} & G & 48909.2 & 1658 & 1127280 & 89177.1 & 1.82 & 6.64 & 1000\\ 
& G+PR & 33109.5 & 1697 & 826931 & 55954.2 & 1.69 & 4.49 & 1000\\ \hline
\multirow{2}{*}{US}& G &  4422351.4 & 5958 & 10138012 & 3719544.2 & 0.84 & 7.76 & 781\\ 
& G+PR & 3789510.5 & 5579 & 10130711 & 3500039.9 & 0.92 & 7.45 & 848\\ \hline
\multirow{2}{*}{N}& G & 3057.6 & 2280 & 6683 & 600.5 & 0.20 & 1.23 & 1000\\ 
& G+PR & 3061.9 & 2123 & 6115 & 585.4 & 0.19 & 1.23 & 1000\\ \hline
\multirow{2}{*}{NS}& G & 3923760.0 & 2681 & 10076751 & 3496925.3 & 0.89 & 7.22 & 845\\ 
& G+PR & 2732969.9 & 13193 & 10014158 & 3048544.5 & 1.11 & 6.81 & 911\\ \hline
\multirow{2}{*}{ND}& G & 2131138.6 & 1633 & 10004863 & 2903255.4 & 1.36 & 10.65 & 929\\ 
& G+PR & 1528595.9 & 1144 & 10002688 & 2548228.5 & 1.67 & 12.00 & 949\\ \hline
\end{tabular}
\caption{Descriptive statistics for the RLDs shown  in Figure~\ref{fig:ggpr_p}; $vc=stddev/mean$ denotes the
variation coefficient, and $q_{0.75}/q_{0.25}$ the quantile ratio, where  $q_x$ denotes the
$x$-quantile.}
\label{tab:ggpr_p}
\end{table}

\begin{figure}
  \centering
  \subfigure[Uniform]{\includegraphics[width = 0.49\textwidth]{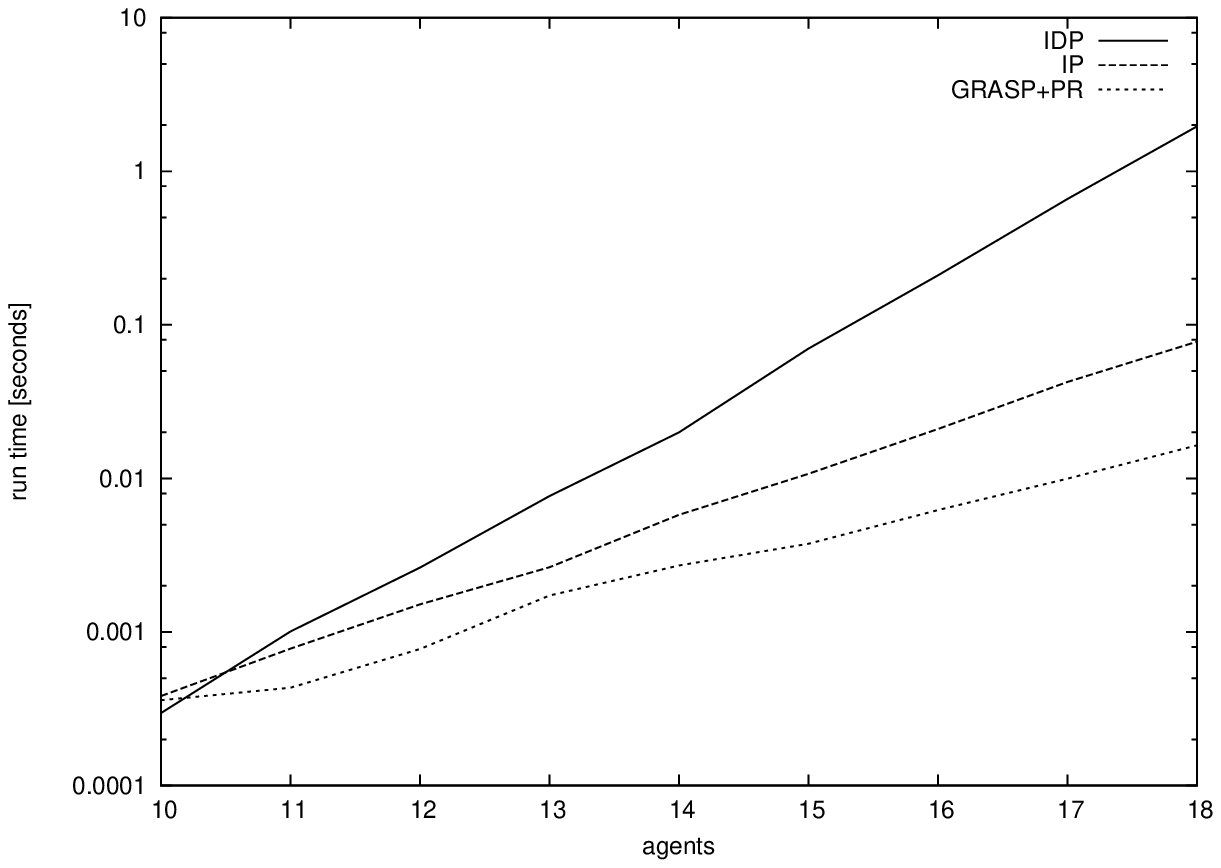}}
  \subfigure[Uniform scaled]{\includegraphics[width = 0.49\textwidth]{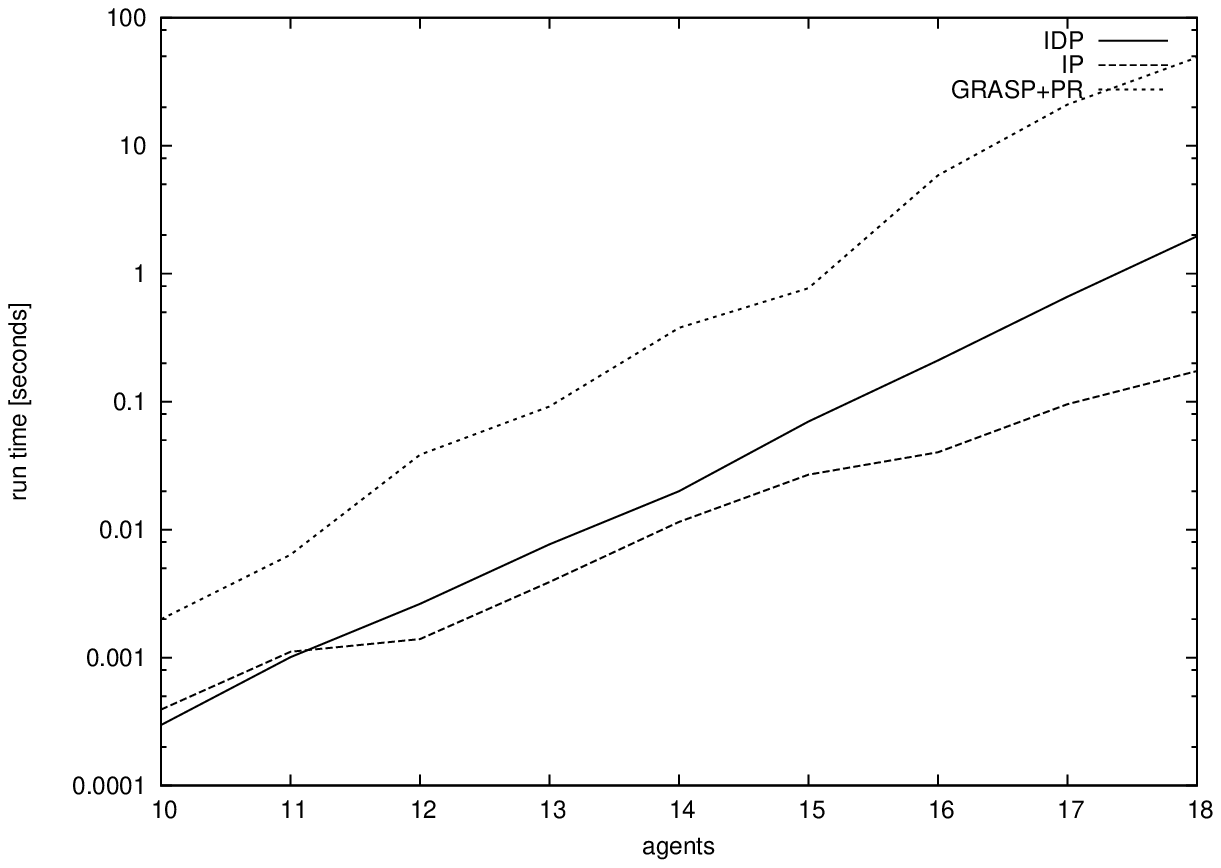}}
  \subfigure[Normal]{\includegraphics[width = 0.49\textwidth]{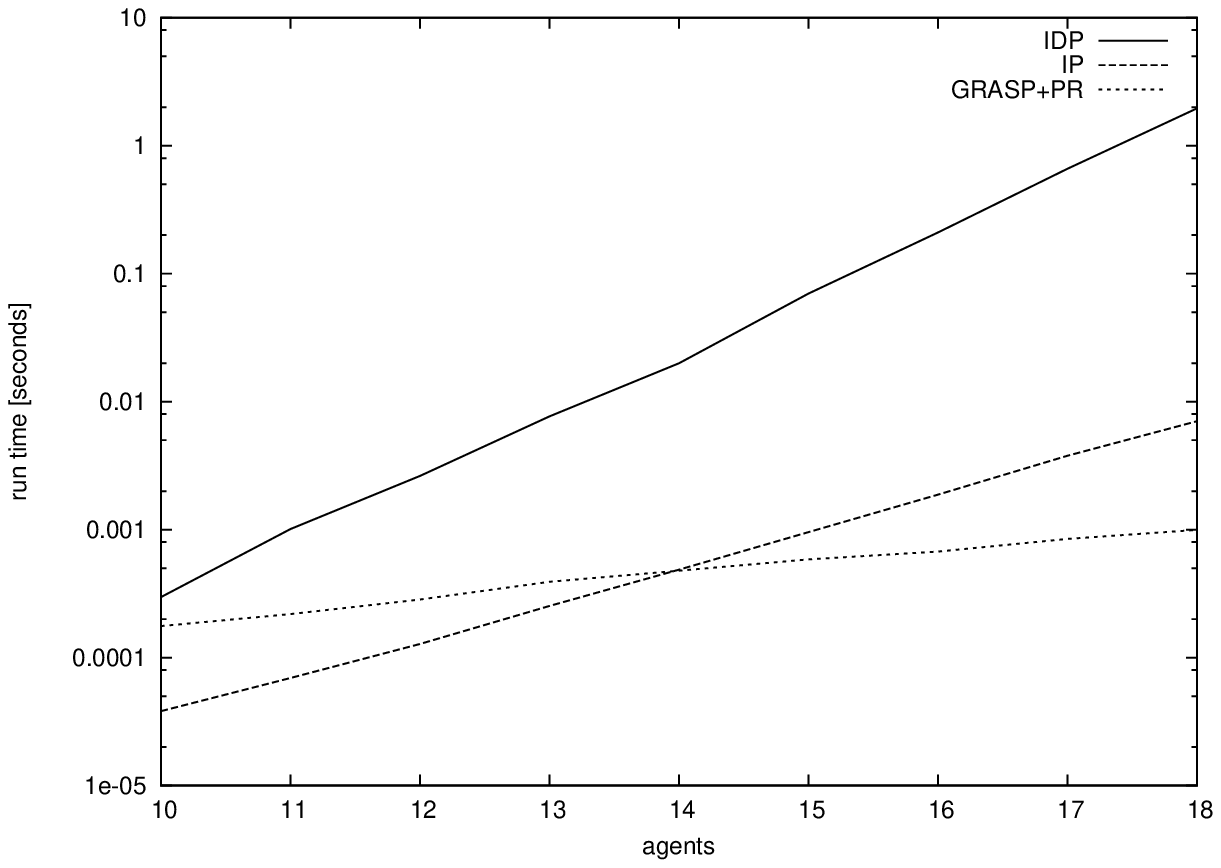}}
  \subfigure[Normal scaled]{\includegraphics[width = 0.49\textwidth]{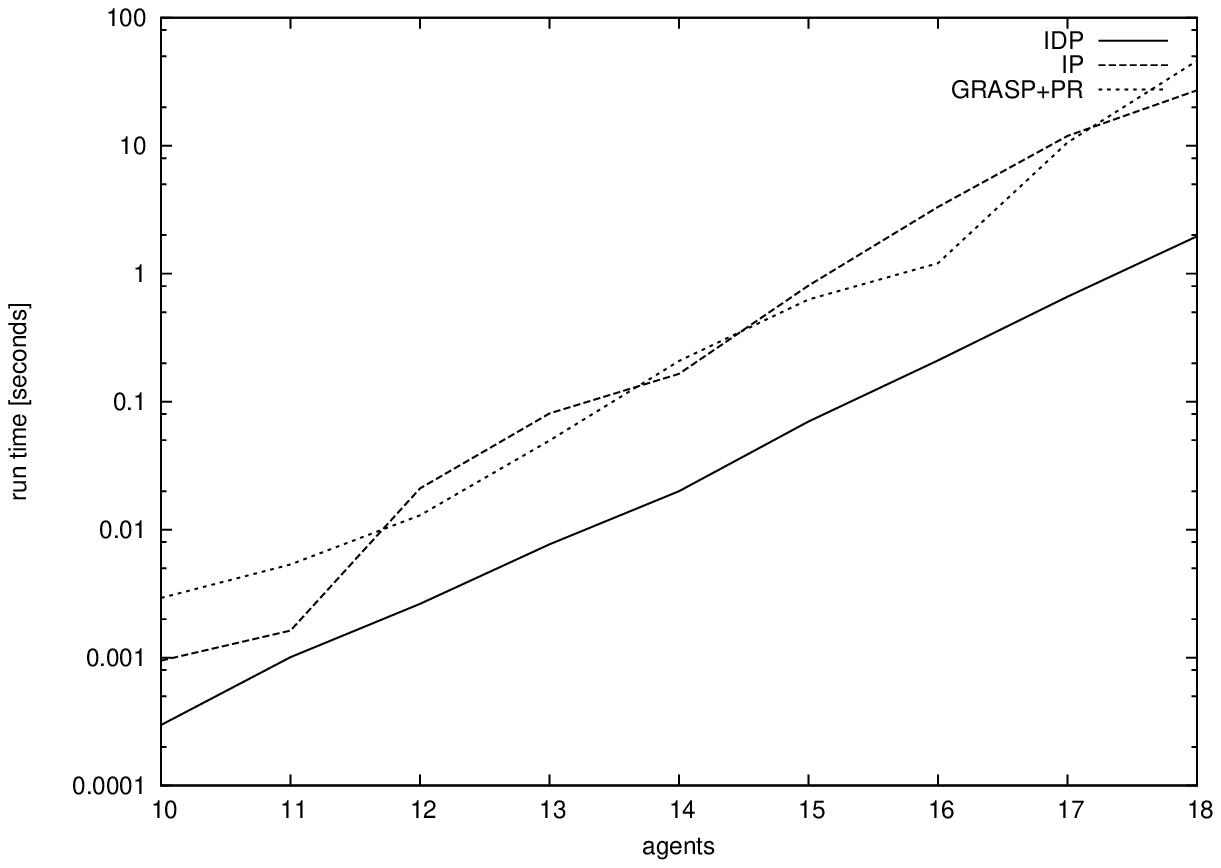}}
  \subfigure[Normal distributed]{\includegraphics[width = 0.49\textwidth]{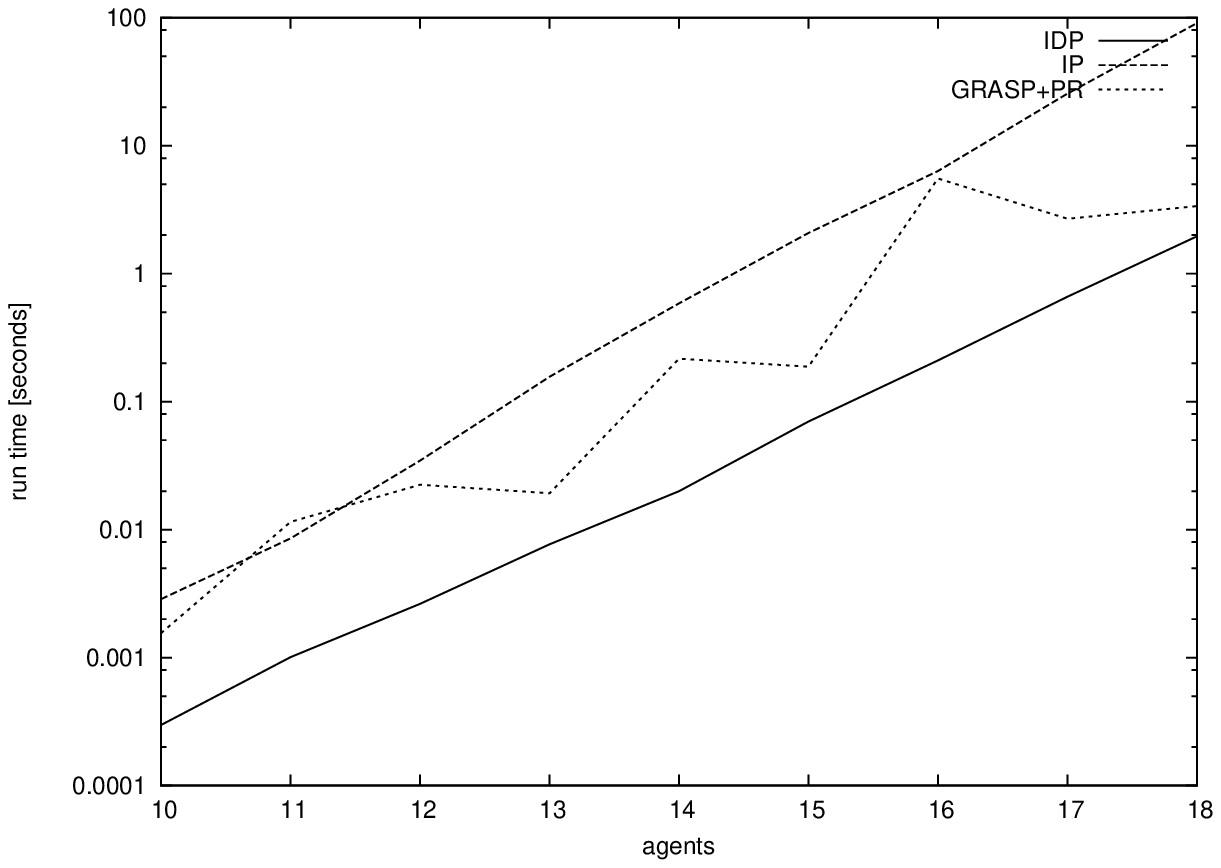}}
  \caption{Semi log-plot of run times for \texttt{GRASP-PR},  with SPLIT/MERGE operator and FORWARD relink
  type, and IP applied to 10 Uniform, Uniform scaled, Normal, Normal scaled, Normal
    distributed CS instances, based on 10 runs per instance.} 
    \label{fig:gip}
\end{figure}

Finally, even if \texttt{GRASP+PR} is not a complete algorithm returning the optimal solution, we
compared it to the two best performing algorithm able to return the optimal solution. 
Given different  numbers of  agents, ranging  from $10$  to $18$, we  compared \texttt{GRASP+PR}  to
\texttt{IDP} and \texttt{IP} reporting  the   time  required   to  find   the  optimal  coalition   structure.  As   reported  in
Figure~\ref{fig:gip},  where  the  time  in  seconds  is plotted  in  a  log  scale,
\texttt{GRASP+PR} outperforms both \texttt{IDP} and \texttt{IP} for non scaled distributions. Its
efficacy is comparable to that of  \texttt{IDP} and \texttt{IP} on the scaled distributions.  

\section{Implementation details}
\label{sec:imp}
\subsection{The characteristic function}
Concerning the representation of the characteristic function and the search space, 
given $n$ agents $N = \{a_1,a_2,\ldots,a_n\}$, we recall that the number of possible coalitions is $2^n-1$.  Hence, the
characteristic function $v : 2^n \rightarrow \mathbb R$ is represented as a vector \texttt{CF} in the following
way. Each subset $S \subseteq A$ (coalition) is described as a binary number $c_{B} = b_1b_2
\cdots b_n $ where each $b_i = 1$ if $a_i \in S$, $b_i = 0$ otherwise. For instance, given $n=4$,
the coalition $\{a_2,a_3\}$  corresponds to the binary number $0110$. Now,  given the binary representation
of a coalition $S$, its decimal value corresponds to the index in the vector \texttt{CF} where 
its corresponding value $v(S)$ is memorised. This gives us the possibility to have a random access to the values
of  the characteristic  functions in  order to  efficiently  compute the  value $v$  of a  coalition
structure. 
\subsection{Coalition structure}
Given a coalition structure $\mathcal C = \{C_1, C_2,  \ldots, C_k\}$, assuming that the $C_i$ are ordered by
their smallest elements, a convenient representation of the CS is an
integer sequence  $d_1 d_2  \cdots  d_n$ where  $d_i =  j$,  if the  agent  $a_i$ belongs  to the  coalition
$C_j$.  Such  sequences  are  known  as  \emph{restricted growth  sequences}~\cite{Milne82}  in  the  combinatorial
literature. The binary representation of the coalition $C_i$ is $b_1 b_2 \cdots b_n$ where $b_j=0$
if $d_j \neq i$, and $b_j=1$ otherwise.

For
instance,  the  sequence  corresponding  to  the  coalition  structure $\mathcal C  =  \{C_1,  C_2,  C_3\}  =
\{\{1,2\},\{3\},\{4\}\}$ is $1123$.    Now  in   order  to  compute   $v(\mathcal C)$,  we   have  to  solve   the  sum
$v(C_1)+v(C_2)+v(C_3)$, where  $C_1$ corresponds to the  binary number $1100$,  $C_2$ corresponds to
the  binary number  $0010$, and  $C_3$ corresponds  to  the binary  number $0001$.   Hence,
$v(\mathcal C)  =
v(C_1)+v(C_2)+v(C_3) =$   \texttt{CF[1100$_2$]}+ \texttt{CF[0010$_2$]}+\texttt{CF[0001$_2$]} = 
\texttt{CF[12]}+\texttt{CF[2]}+\texttt{CF[1]}, where \texttt{CF} is  the vector containing the  values of the
characteristic function.

\section{Conclusions}
\label{sec:conc}
The paper presented an algorithm applyable to cooperative
complex problems that require to find an optimal partition, maximising a social welfare, of a set of
entities involved  in a system  into exhaustive  and disjoint coalitions.  We present  a greedy  adaptive search
procedure  with  path-relinking  to efficiently  search the  space of  coalition structures of those
gouping problems.
As reported  in the experimental  section the proposed  algorithm
outperforms in some cases the state of the art algorithms in computing optimal coalition structures.

\bibliographystyle{amsplain}
\bibliography{coalition}






% \end{thebibliography}

\end{document}